\documentclass{article}

\usepackage{microtype}
\usepackage{graphicx}
\usepackage{subcaption}
\usepackage{booktabs} 
\usepackage{multirow}

\usepackage{hyperref}

\usepackage[preprint]{icml2026}

\usepackage{amsmath}
\usepackage{enumitem}
\usepackage{amssymb}
\usepackage{mathtools}
\usepackage{amsthm}
\usepackage{algorithm}
\usepackage{algorithmic}

\usepackage[capitalize,noabbrev]{cleveref}

\theoremstyle{plain}
\newtheorem{theorem}{Theorem}[section]
\newtheorem{proposition}[theorem]{Proposition}
\newtheorem{lemma}[theorem]{Lemma}

\theoremstyle{definition}

\newtheorem{assumption}[theorem]{Assumption}
\crefname{assumption}{Assumption}{Assumptions}
\crefname{appendix}{Appendix}{Appendices}
\Crefname{appendix}{Appendix}{Appendices}
\Crefname{assumption}{Assumption}{Assumptions}
\theoremstyle{remark}
\newtheorem{remark}[theorem]{Remark}

\newcommand{\E}{\mathbb{E}}
\newcommand{\Prob}{\mathbb{P}}
\newcommand{\cS}{\mathcal{S}}
\newcommand{\cA}{\mathcal{A}}
\newcommand{\cM}{\mathcal{M}}
\newcommand{\cF}{\mathcal{F}}
\newcommand{\cR}{\mathcal{R}}
\newcommand{\cC}{\mathcal{C}}
\newcommand{\cand}{\tilde{\pi}}
\newcommand{\simT}{\tilde{T}}

\newcommand{\gc}{\gamma_c}
\newcommand{\Wmax}{\rho_{\max}}
\newcommand{\cmax}{c_{\max}}

\newcommand{\Nmax}{N_{\max}}
\newcommand{\cW}{\rho_{0:t}}
\newcommand{\cWc}{\bar{\rho}_{0:t}}

\usepackage[textsize=tiny]{todonotes}

\icmltitlerunning{Dejavu}

\begin{document}

\twocolumn[
  \icmltitle{Probably Safe Sim-to-Real Transfer with Lifetime Validity}

  \icmlsetsymbol{equal}{*}

  \begin{icmlauthorlist}
    \icmlauthor{Rohan Tangri}{ox}
    \icmlauthor{Jan-Peter Calliess}{ox}
  \end{icmlauthorlist}

  \icmlaffiliation{ox}{Machine Learning Research Group, University of Oxford, Oxford, United Kingdom}

  \icmlcorrespondingauthor{Rohan Tangri}{rohan.tangri@reub.ox.ac.uk}
  \icmlcorrespondingauthor{Jan-Peter Calliess}{jan@robots.ox.ac.uk}

  \icmlkeywords{safe reinforcement learning, sim-to-real, off-policy evaluation, confidence sequences}

  \vskip 0.3in
]

\printAffiliationsAndNotice{}  

\begin{abstract}
  We present Dejavu, a method that makes sim-to-real transfer of reinforcement learning policies in constrained Markov decision processes \emph{probably safe} over an unbounded sequence of policy updates. An anytime valid test using per-decision importance sampling and a betting confidence sequence certifies each candidate trained in simulation before deployment using only trajectories collected by a safe baseline policy operating in production. Simulation quality does not affect the safety guarantee, only the production sample efficiency of the method. Finally, a summable per-test failure probability yields the lifetime valid property. Empirical results across three sim-to-real benchmarks support the theoretical work.
\end{abstract}

\section{Introduction}
\label{sec:intro}

\emph{Reinforcement learning} (RL) policies in high-stakes systems are rarely trained where they will be used \citep{tobin2017domain, peng2018sim}. Training in a simulation environment is preferred for at least two reasons. First, \emph{safe trial and error}: a learner in simulation can test unsafe policies with no real risk, whereas exploration in the production environment must visit harmful states to learn to avoid them \citep{garcia2015comprehensive}. Second, \emph{practical timelines}: simulated experience is often orders of magnitude cheaper and faster than production experience \citep{rudin2022learning}.

The price is the \emph{sim-to-real gap}. A policy that satisfies its safety constraint in the simulation environment can violate it in production, which is dangerous exactly when safety must hold on first contact. Domain randomization exposes the policy to a distribution over simulation parameters aiming to encompass the production environment setting \citep{tobin2017domain, peng2018sim}, and its pessimistic refinement SPiDR penalizes visits to states where randomized models disagree \citep{as2025spidr}. These methods shrink the gap, but they do not certify whether deployment is safe for the specific policy in hand and the specific production environment in front of it. In current practice, that decision rests on unquantified judgment or on holdout trials of the very policy whose safety is in doubt.

This paper provides the missing decision procedure. Our setting includes a \emph{baseline policy} operating safely in the \emph{production environment} (an incumbent controller, a previously validated policy) and a \emph{candidate policy} which has been trained as its replacement in the \emph{simulation environment}. The question is: \emph{can policies trained in simulation be deployed to production over an unbounded sequence of updates, such that with high probability every deployed policy is safe, however inaccurate the simulator?}

Our answer is Dejavu,\footnote{The name describes the mechanism: every test is computed entirely from trajectories of the currently deployed baseline policy. The candidate is evaluated on d\'ej\`a vu, before it ever acts.} which offers the following contributions:
\begin{itemize}[nosep, itemsep=2pt]
  \item \textbf{An anytime-valid test}
  (\cref{sec:certificate}): An upper confidence bound on a candidate's production-environment cost return. The bound is valid at every buffer size at once, so the test condition can be checked as production data arrives and stop the moment it passes.
  \item \textbf{Lifetime validity}
  (\cref{sec:ladder}): Composing tests into a sequence, the probability that any policy ever deployed violates the cost limit is bounded by a summable budget in a sample efficient manner.
  \item \textbf{Data cost and simulation quality} (\cref{sec:rates}): Lower and upper bounds on the expected production data needed to pass a test, and an analysis of how simulation quality affects that data cost.
  \item \textbf{Empirical validation} (\cref{sec:experiments}):
  Experiments across three sim-to-real benchmarks with deliberately miscalibrated simulation environments.
\end{itemize}

\section{Related Work}
\label{sec:related}

\paragraph{Safe sim-to-real transfer.}
Domain randomization is the de facto tool for crossing the reality gap \citep{tobin2017domain, peng2018sim}, but offers no safety guarantee at deployment. SPiDR \citep{as2025spidr} augments the training cost with a pessimism term proportional to the disagreement among randomized dynamics models, inheriting guarantees with respect to the assumed uncertainty set. The guarantee attaches to the training procedure, not to the individual trained policy in the actual production environment. Robust MDPs \citep{iyengar2005robust} and their constrained extensions \citep{queeney2023risk, kitamura2024epigraph} similarly optimize against an uncertainty set over dynamics. Sim-to-Lab-to-Real \citep{hsu2023sim} interposes a high-fidelity intermediate environment and a learned shield, while \citet{kaushik2022safeapt} select among policies trained in simulation with safe Bayesian optimization on production data. Dejavu is complementary to this entire family: any of these methods can serve as the proposer of candidates, while Dejavu supplies a per-policy, per-environment deployment test computed before the candidate acts, from data whose collection incurs no new risk.

\paragraph{High-confidence policy improvement.}
High-confidence off-policy evaluation bounds a candidate's return from a baseline policy's trajectories \citep{thomas2015high}, and high-confidence policy improvement (HCPI) deploys a candidate only if this bound clears a threshold, which \citet{thomas2015improvement} call a \emph{safety test}. Seldonian algorithms extend the test to general constraints \citep{thomas2019preventing}, and batch safe policy improvement bounds the loss relative to a baseline \citep{laroche2019safe}. These tests fix their amount of data in advance; their exact variants are conservative and their practical variants approximate. Their guarantees also cover one test at a time: HCPI's iterative version reuses data across refused candidates, as Dejavu does, but its probability of some unsafe deployment grows with the number of tests, and the version that also reuses data across deployments keeps its guarantee only until the first deployment \citep{thomas2015improvement}. Dejavu keeps one guarantee over an unlimited sequence of deployments while reusing data, by splitting the allowed probability of an unsafe deployment across its tests and handing a passing test's share on to later tests \citep{tian2021online}, as \citet{feng2022sequential} do for model updates outside RL. Unlike these methods, Dejavu tests candidates trained in simulation. \citet{katdare2023marginalized} also combine simulation and production data for evaluation, but must learn density ratios, while Dejavu's weights are computed.

\paragraph{Anytime-valid tests.}
Anytime-valid confidence sequences hold at every sample size, so a test built on them can stop as soon as the data suffice \citep{howard2021time, waudby2024estimating, shekhar2023near}. \citet{karampatziakis2021off} use them to decide deployment in contextual bandits, and \citet{waudby2024anytime} for off-policy evaluation, both for one-step decisions. Dejavu applies them to multi-step tasks; the sequence of tests in \cref{sec:ladder} holds for any anytime-valid test.

\section{Preliminaries}
\label{sec:prelim}

\paragraph{Constrained Markov Decision Process (CMDP), real and simulated.}
Let $\Delta(\mathcal{X})$ represent
a set of probability distributions over a set $\mathcal{X}$. The production environment is a constrained Markov decision process (CMDP; \citealp{altman1999constrained}) $\cM = (\cS, \cA, T, \cR, \cC, d_0)$. Here, $\cS$ is the state space, $\cA$ is the action space, $T: \mathcal{S} \times
\mathcal{A} \rightarrow \Delta(\mathcal{S})$ is the transition kernel, $\cR : \cS \times \cA \to \Delta(\mathbb{R})$ is the reward kernel, $\cC : \cS \times \cA \to \Delta([0, \cmax])$ is the cost kernel and $d_0 \in \Delta(\mathcal{S})$ is the initial state distribution. A policy $\pi: \mathcal{S} \rightarrow \Delta(\mathcal{A})$
maps states to a probability distribution over actions. We assume the agent
interacts with the environment to generate infinite length \emph{trajectories} $\tau =
(s_t, a_t, r_t, c_t)_{t \ge 0}$, which are sampled at time $t$ with \emph{actions} $a_t \sim \pi(\cdot|s_t)$, \emph{states} $s_{t+1} \sim T(\cdot|s_t, a_t)$, \emph{rewards} $r_t \sim \cR(\cdot|s_t, a_t)$ and \emph{costs} $c_t \sim \cC(\cdot|s_t, a_t)$. The simulation environment is a CMDP $\tilde{\cM} = (\cS, \cA, \simT, \cR, \cC, \tilde d_0)$ differing only in state-transition dynamics and initial-state distribution. Throughout, a tilde marks simulation objects. 

For a policy $\pi$ define the \emph{reward return}, $G(\tau) = \sum_{t \ge 0} \gamma^t r_t$, and \emph{cost return}, $C(\tau) = \sum_{t \ge 0} \gc^t c_t$, with their expectations in the production environment, $J(\pi) = \E_{\tau \sim (\pi, \cM)}[G(\tau)]$ and $J_C(\pi) = \E_{\tau \sim (\pi, \cM)}[C(\tau)]$. Here, $\gamma, \gc \in [0, 1)$ are the \emph{reward and cost discount factors}, and $\tau \sim (\pi, \cM)$ denotes a trajectory sampled by policy $\pi$ in the production environment. $\tilde{J}(\pi)$ and $\tilde{J}_C(\pi)$ are their simulation-environment counterparts, with trajectories sampled $\tau \sim (\pi, \tilde\cM)$. The design goal is the standard CMDP program \citep{achiam2017constrained, ray2019benchmarking}
\begin{equation}
  \label{eq:goal}
  \boxed{\pi^\star = \arg\max_\pi \; J(\pi) \quad \text{s.t.} \quad J_C(\pi) \le \ell }
\end{equation}
for a fixed \emph{cost limit} $\ell$. A policy is \emph{safe} if it satisfies the cost constraint, $J_C(\pi) \le \ell$, and \emph{unsafe} otherwise.

\paragraph{Per-decision importance sampling (PDIS).}
We are forbidden from sampling trajectories in the production environment using the candidate before it is certified: $\tau \sim (\cand, \cM)$. Per-decision importance sampling (PDIS) allows us to evaluate expectations over such trajectories from production environment trajectories sampled with the baseline policy: $\tau \sim (\pi, \cM)$. For a candidate $\cand$ and the baseline policy $\pi$, define the \emph{per-step importance weight}
\begin{equation}
  \label{eq:per-step-is}
  \rho(s, a) = \frac{\cand(a \mid s)}{\pi(a \mid s)} \ge 0,
\end{equation}
and the \emph{cumulative importance weight}
\begin{equation}
  \label{eq:cum-is}
  \cW(\tau) = \prod_{i=0}^{t} \rho(s_i, a_i) .
\end{equation}
Let the \emph{importance-weighted cost return} be
\begin{equation}
  \label{eq:importance-weighted-cost-return}
  C_\rho(\tau) = \sum_{t \ge 0} \gc^t\, \cW(\tau)\, c_t,
\end{equation}
then per-decision importance sampling (PDIS; \citealp{precup2000eligibility}) makes the following equivalence:
\begin{equation}
\label{eq:pdis}
  J_C(\cand) = \E_{\tau \sim (\cand, \cM)}\Big[C(\tau)\Big] = \E_{\tau \sim (\pi, \cM)}\Big[C_\rho(\tau)\,\Big].
\end{equation}

\paragraph{Betting confidence sequences.}
Let $Y_1, Y_2, \ldots \in [0,1]$ be i.i.d.\ samples with unknown mean $\mu$. We aim to achieve an \emph{anytime valid} upper bound on $\mu$: $\Prob(\exists n : \mu > u_n) \le \alpha$, which holds at any data-dependent stopping time unlike fixed-$n$ concentration bounds (e.g. empirical-Bernstein \citep{maurer2009empirical}). Ville's inequality gives such guarantees \citep{ville1939etude}: if $K_0 = 1, K_1, K_2, \ldots$ is a nonnegative supermartingale, meaning $K_n \ge 0$ and $\E[K_{n+1} \mid \cF_n] \le K_n$ for the history $\cF_n$ of the first $n$ samples, then $\Prob(\exists n : K_n \ge 1/\alpha) \le \alpha$.

Following \citet{waudby2024estimating}, for a hypothesized mean $m \in [0,1]$ and $K_0(m) = 1$, define the \emph{capital process}
\begin{equation}
  \label{eq:capital}
  K_n(m) = K_{n-1}(m)\,\big(1 + \lambda_n(m) (m - Y_n)\big),
\end{equation}
where $\lambda_n(m) \in [0, \tfrac{1}{1-m}]$, such that $K_n(m) \ge 0$, and is \emph{predictable}, a function of $m$ and $Y_1, \ldots, Y_{n-1}$ only. Construct an upper bound as
\begin{equation}
  \label{eq:ucs}
  u_n = \sup\{ m \in [0,1] : K_n(m) < 1/\alpha \}.
\end{equation}
Since $K_n(m)$ is positive, continuous and nondecreasing in $m$ under the variance-adaptive plug-in schedule of \citet{waudby2024estimating} used by Dejavu (\cref{app:betting}), $K_n(u_n) = 1/\alpha$ when $u_n < 1$, and then:
\begin{equation}
  \label{eq:equivalence}
  m \ge u_n \Longleftrightarrow K_n(m) \ge K_n(u_n) \Longleftrightarrow K_n(m) \ge \frac{1}{\alpha}.
\end{equation}
When $m = \mu$, $K_n(\mu)$ is a nonnegative supermartingale started at $1$, so Ville's inequality gives $\Prob(\exists n : K_n(\mu) \ge 1/\alpha) \le \alpha$. For $\mu < 1$, the event $\mu \ge u_n$ has $u_n < 1$, so by \cref{eq:equivalence}, $\mu \ge u_n \Longleftrightarrow K_n(\mu) \ge 1 /\alpha$, and $\Prob(\exists n : \mu \ge u_n) \le \alpha$. Since $\{\mu > u_n\} \subseteq \{\mu \ge u_n\}$, we obtain $\Prob(\exists n : \mu > u_n) \le \alpha$ \citep{howard2021time, waudby2024estimating}. If $\mu = 1$, then $\mu = u_n$ and the bound never fails.

\section{Dejavu's Test}
\label{sec:certificate}

We first specify one \emph{test}, indexed by $k \in \mathbb{N}_0$, to promote or refuse a candidate $\cand_k$ using a \emph{buffer} $D_k$ of trajectories with finite horizon $H$ sampled from a safe baseline policy in the production environment, $(\pi_{k}, \cM)$. The sequence of tests is \cref{sec:ladder}. A \emph{proposer} trains the candidate in the simulation environment without any data from the buffer and warm starting from the baseline policy. The test is agnostic to the proposer, but an example is given in the optimization problem (\emph{OP}),
\begin{equation}
  \label{eq:proposer}
  \begin{aligned}
  \cand_k = \arg\max_{\tilde \pi} \;& \tilde{J}(\tilde \pi) - \beta\, \bar D_{\mathrm{KL}}(\tilde \pi \,\|\, \pi_k) \\
  \text{s.t.} \;& \tilde{J}_C(\tilde \pi) \le \tilde{\ell}.
  \end{aligned}
\end{equation}
Here, $\bar D_{\mathrm{KL}}(\tilde \pi \,\|\, \pi_k) = \E_{s \sim (\tilde \pi, \tilde\cM)}\big[D_{\mathrm{KL}}\big(\tilde \pi \,\|\, \pi_k\big)\big]$ is the KL divergence \emph{anchor}, $\tilde{\ell}$ is a \emph{simulation cost limit} that need not equal $\ell$, since the simulation environment may be miscalibrated, and $\beta \ge 0$ is the \emph{anchor coefficient}. The anchor is not required by the guarantee, but it makes \cref{ass:cap} hold in practice (\cref{app:perstep}) and it reduces the production data a test needs (\cref{sec:rates}).

\begin{assumption}[Bounded costs]
\label[assumption]{ass:bounded}
$c_t \in [0, \cmax]$ almost surely.
\end{assumption}

\begin{assumption}[Cumulative importance weight clip]
\label[assumption]{ass:cap}
There is a known $\Wmax > 0$ such that, almost surely over $\tau \sim (\pi_{k}, \cM)$, $\cW(\tau) \le \Wmax$ for all $t < H$.
\end{assumption}

\begin{assumption}[Candidate--buffer independence]
\label[assumption]{ass:indep}
The candidate $\cand_k$ is never trained or chosen using production data or test outcomes from the current baseline policy $\pi_k$.
\end{assumption}

\begin{assumption}[Safe start]
\label[assumption]{ass:start}
The initial policy is safe, $J_C(\pi_0) \le \ell$, either by construction or verified at confidence $1 - \alpha_{\pi_0}$ by applying the betting confidence sequence of \cref{sec:prelim} to the discounted cost returns of $\pi_0$'s own production trajectories.
\end{assumption}

The true value of $J_C(\cand_k)$ (\ref{eq:pdis}) is unknown, so we instead aim to calculate an anytime-valid upper confidence bound using the sampled buffer of production trajectories with a betting confidence sequence. This requires bounded samples of the importance-weighted cost return (\ref{eq:importance-weighted-cost-return}). However, in practice, we only collect finite horizon trajectories of length $H$, and \cref{ass:cap} can be violated, with unbounded cumulative importance weights. Therefore, we define the per-trajectory \emph{statistic}
\begin{equation}
\label{eq:certvalue}
  X(\tau) = \sum_{t<H} \gc^t\, \cWc(\tau)\, c_t \in [0, w_X],
\end{equation}
where $\cWc(\tau) = \min\big(\cW(\tau), \Wmax\big)$ is the \emph{clipped cumulative importance weight} and the range has width $w_X \coloneqq \Wmax\, \cmax\, \frac{1 - \gc^H}{1 - \gc}$. The worst-case horizon truncation error is the \emph{truncation offset} $c_{>H} \;\coloneqq\; \tfrac{\gc^H\, \cmax}{1 - \gc}$.
\begin{lemma}[Cost return bound]
\label[lemma]{lem:envelope}
Under \cref{ass:bounded,ass:cap}, for a fixed candidate $\cand_k$, the statistic of \cref{eq:certvalue} satisfies (proof \cref{app:envelope}):
\begin{equation}
  J_C(\cand_k) \;\le\; \E_{\tau \sim (\pi_k, \cM)}\big[X(\tau)\big] + c_{>H}
\end{equation}
\end{lemma}

The clip is what makes the statistic bounded, and \cref{ass:cap} asserts that it never binds. However, if the assumption fails, the clip lowers $X(\tau)$ below its unclipped counterpart at some trajectories, so the right-hand side of \cref{lem:envelope} is biased downward, the only anti-conservative direction in the statistic. Therefore, a clipping method that minimizes the bias introduced is preferred. For a candidate $\tilde \pi$ let its truncated cost return be $J^H_C(\tilde \pi) = \E_{\tau \sim (\tilde \pi, \cM)}[\sum_{t<H} \gc^t c_t]$.
\begin{proposition}[Cumulative clip bias]
\label[proposition]{prop:clip}
Under \cref{ass:bounded}, with $\pi_k(a \mid s) > 0$ wherever $\cand_k(a \mid s) > 0$, and with expectations over $\tau \sim (\pi_k, \cM)$ (proof \cref{app:clipproof}),
\begin{equation}
  0 \;\le\; J^H_C(\cand_k) - \E[X(\tau)] \;\le\; \cmax \sum_{t<H} \gc^t (1 - \varrho_t),
\end{equation}
where $\varrho_t \coloneqq \E[\cWc(\tau)] \le 1$ is the \emph{retained mass} at step $t$, and $\varrho_t = 1$ under \cref{ass:cap}.
\end{proposition}

\cref{prop:clip} gives an upper bound on the cumulative clip bias that is zero under \cref{ass:cap}. The retained mass is the observable audit of \cref{ass:cap}, one when it holds and less than one otherwise. The natural alternative is to clip each per-step importance weight before forming the product, as in Retrace \citep{munos2016safe} and V-trace \citep{espeholt2018impala}. Since $\E_{a \sim \pi_k(\cdot \mid s)}[\rho(s, a)] = 1$ at every state, per-step weights above and below one offset inside the product, and the cumulative clip acts only on what remains, while the per-step clip acts on each factor before any offsetting can occur. \cref{app:perstep} shows that its bias consequently has a lower bound that grows with the candidate's excess cost over the baseline policy, which is why Dejavu clips the cumulative importance weight.

Given $n$ trajectories indexed by $j$: $\{\tau_1, \dots, \tau_n\} \sim (\pi_k, \cM)$. Let $Y(\tau) = X(\tau) / w_X \in [0, 1]$ and run the betting confidence sequence with the capital process of \cref{eq:capital} at \emph{test level} $\alpha_k$. Rescaling the normalized upper bound in \cref{eq:ucs} by $w_X$ yields the upper bound for $\E_{\tau \sim (\pi_k, \cM)}\big[X(\tau)\big]$:
\begin{equation}
  \label{eq:Un}
  U_n(\cand_k) = w_X\, u_n.
\end{equation}
The test promotes $\cand_k$ if
\begin{equation}
  \label{eq:gate}
  \boxed{ U_n(\cand_k) + c_{>H} \;\le\; \ell . }
\end{equation}
We call \cref{eq:gate} the \emph{test condition}. It is checked repeatedly as the buffer grows by $n_0$ trajectories at a time. The test stops at \emph{promotion time} $N_P = \min\{n : U_n(\cand_k) + c_{>H} \le \ell\}$, or at a data budget $\Nmax$ if it never does (\cref{alg:dejavu}). The random number of production trajectories the test consumes is the \emph{stopping time}
\begin{equation}
  \label{eq:stopping}
  N_k \coloneqq \min(N_P, \Nmax).
\end{equation}

\begin{theorem}[Validity of each test]
\label{thm:rung}
Fix test $k$ and condition on its past $\cF_{<k}$ (all data, candidates, and decisions before test $k$). Under \cref{ass:bounded,ass:cap,ass:indep}, for the stopping time $N_k$ of \cref{eq:stopping},
\begin{equation}
  \Prob\big( J_C(\cand_k) > U_{N_k}(\cand_k) + c_{>H} \,\big|\, \cF_{<k} \big)
  \;\le\; \alpha_k .
\end{equation}
See \cref{app:rung} for a proof.
\end{theorem}

\begin{remark}[Bound failure is not safety failure]
\label{rem:calibration}
The test level $\alpha_k$ bounds the probability of a \emph{bound failure}, $J_C(\cand_k) > U_{N_k}(\cand_k) + c_{>H}$. A false promotion requires $J_C(\cand_k) > \ell \ge U_{N_k}(\cand_k) + c_{>H}$ and is therefore a bound failure, so it occurs with probability at most $\alpha_k$. The converse does not hold: a bound failure on a refused candidate is harmless, and one on a promoted candidate is harmless whenever $J_C(\cand_k) \le \ell$. False promotions are therefore at least as rare as bound failures.
\end{remark}

\section{A Sequence of Tests}
\label{sec:ladder}

A deployed system does not certify once. It may run the test of \cref{sec:certificate} indefinitely: train, freeze, certify, promote or refuse, re-anchor, repeat (\cref{alg:dejavu}). Each test's guarantee has a false-promotion probability of at most its test level $\alpha_k$, and a union bound over tests adds these. A constant level makes the sum exceed one after finitely many tests, so for a sequence of tests that must run indefinitely the lifetime guarantee is vacuous. We therefore fix a \emph{lifetime budget} $\alpha \in (0, 1)$ and release it through the \emph{alpha-spending} schedule of online multiple testing \citep{tian2021online},
\begin{equation}
  \label{eq:budget}
  \eta_k = \frac{1}{(k+1)(k+2)}
  \quad \Longrightarrow \quad
  \sum_{k=0}^{\infty} \eta_k\, \alpha = \alpha ,
\end{equation}
where $\eta_k \alpha$ is the \emph{injection} released at the $k$-th release point. The simplest sequence uses test level $\alpha_k = \eta_k \alpha$ on a fresh buffer and discards both the level and buffer when the test ends. This is wasteful on production data. Instead, Dejavu first reuses the buffer when a test refuses, so $D_{k+1} \leftarrow D_k$, and more production data is only collected if $\cand_{k+1}$ doesn't pass on the inherited buffer. Second, Dejavu passes the level to the next test when it promotes using the \emph{online fallback} procedure of \citet{tian2021online} (\cref{alg:dejavu}). The tests are therefore grouped into \emph{streams}, each of which runs one or more tests with the same baseline policy.

Let $i_k \in \mathbb{N}_0$ be the stream that test $k$ belongs to, and $p_k \in \mathbb{N}_0$ its position in the stream. Injections $\eta_i\alpha$ are assigned to each stream $i$. The number of tests in a stream depends on the buffer that will be collected. However, \cref{thm:rung} requires the test level to be fixed independently of the buffer it is tested on. Dejavu fixes the levels of a stream when it starts, over every test the stream could run. This leaves a trade-off. Giving most of the stream's level to its first test leaves the later tests too little level to promote from. Holding most of the stream's level for the later tests costs those later levels when the first test promotes. A \emph{reserve} $\phi_i \in (0,1)$ defines the fraction of the stream injection for the test levels after the first. The test levels in a stream are set by extending the online fallback of \citet{tian2021online}:
\begin{equation}
  \label{eq:levels}
    \alpha_k =
    \begin{cases}
      \alpha_{k-1} + (1 - \phi_{i_k})\, \eta_{i_k}\,\alpha, & p_k = 0, \\[3pt]
      \dfrac{\phi_{i_k}\, \eta_{i_k}\,\alpha}{p_k (p_k + 1)}, & p_k \ge 1,
    \end{cases}
\end{equation}
where $\alpha_{-1}=0$. \cref{thm:lifetime} allows $\phi_i$ to be any function of what was observed before stream $i$ started. Given test $k$ at the start of stream $i$, Dejavu uses
\begin{equation}
  \label{eq:reserve}
  \phi_i = \frac{k - i + 1}{k + 2}.
\end{equation}
On a promotion streak $\phi_i \to 0$, each test passes on almost all it had, and the test level tends to $\alpha \sum_{i \ge 0} (1 - \phi_i)\,\eta_i = (\pi^2/6 - 1)\,\alpha \approx 0.65\,\alpha$ under \cref{eq:budget} instead of decaying to zero. Where refusals are common, $\phi_i$ approaches the refusal rate. A later test that promotes passes its own level, its share of the reserve, to the first test of the next stream.

\begin{algorithm}[t]
  \caption{Dejavu}
  \label{alg:dejavu}
  \begin{algorithmic}
    \STATE {\bfseries Input:} initial policy $\pi_0$ (safety given or verified at level $\alpha_{\pi_0}$; \cref{ass:start}); cost limit $\ell$; lifetime budget $\alpha$; clip $\Wmax$; horizon $H$; chunk size $n_0$; per-test data budget $\Nmax$; early-refusal level $\beta$ (\cref{eq:early-refusal}).
    \STATE $D_0 \gets \emptyset$; $i \gets 0$; $p \gets 0$
    \FOR{$k = 0, 1, \ldots$}
      \STATE If $p = 0$, set $\phi_i$ by \cref{eq:reserve} \COMMENT{start of stream $i$}
      \STATE Set $\alpha_k$ by \cref{eq:levels}
      \STATE Train $\cand_k$ in the simulation environment from $\pi_k$ (e.g.\ \cref{eq:proposer}); \textbf{freeze} it
      \STATE Score $D_k$ under $\cand_k$ by \cref{eq:certvalue}; compute $U_n(\cand_k)$ at level $\alpha_k$ on its $n$ trajectories
      \WHILE{$U_n(\cand_k) + c_{>H} > \ell$, \cref{eq:early-refusal} does not hold, and fewer than $\Nmax$ trajectories were added at this test}
        \STATE Collect $n_0$ trajectories of $(\pi_k, \cM)$, append them to $D_k$, and update $U_n(\cand_k)$
      \ENDWHILE
      \IF{$U_n(\cand_k) + c_{>H} \le \ell$}
        \STATE \textbf{promote:} $\pi_{k+1} \gets \cand_k$; $D_{k+1} \gets \emptyset$; $i \gets i + 1$; $p \gets 0$
      \ELSE
        \STATE \textbf{refuse:} $\pi_{k+1} \gets \pi_k$; $D_{k+1} \gets D_k$; $p \gets p + 1$
      \ENDIF
    \ENDFOR
  \end{algorithmic}
\end{algorithm}

\begin{theorem}[Lifetime validity of Dejavu]
\label{thm:lifetime}
Under \cref{ass:indep,ass:start}, suppose that a test satisfies $\Prob\big(\exists n : J_C(\cand) > U_n(\cand) + c_{>H}\big) \le \alpha'$ for any candidate $\cand$ and level $\alpha'$ fixed before a buffer of i.i.d.\ trajectories of the baseline policy is collected. Dejavu's test satisfies this under \cref{ass:bounded,ass:cap}. Run \cref{alg:dejavu} for any number of such tests, and let $\alpha_{\pi_0}$ be the bound on the probability that $\pi_0$ is unsafe (\cref{ass:start}). Then
\begin{equation}
  \Prob\big( \exists\, k : J_C(\pi_k) > \ell \big) \;\le\; \alpha_{\pi_0} + \alpha .
\end{equation}
See \cref{app:lifetimeproof} for a proof.
\end{theorem}

The two terms cover the two failure modes: some test false-promotes (at most $\alpha$) or the initial policy was never safe (at most $\alpha_{\pi_0}$). Any schedule that has test levels fixed at the start of each stream and where the sum of the test levels in any stream is at most the stream injection plus the level passed to it satisfies \cref{thm:lifetime}.
\section{Data Cost and Simulation Quality}
\label{sec:rates}
The \emph{data cost} of test $k$ is $\E[N_k]$, the expected number of production trajectories it consumes before the stopping time of \cref{eq:stopping}. It is governed by the \emph{slack} 
\begin{equation}
  \label{eq:slack}
  s_k \coloneqq \ell - c_{>H} - \E_{\tau \sim (\pi_k, \cM)}[X(\tau)].
\end{equation}
The test promotes a candidate when the test condition \cref{eq:gate} holds, which, with the \emph{normalized threshold} $m_\ell = (\ell - c_{>H})/w_X$ and \cref{eq:equivalence}, is equivalent to
\begin{equation}
  \label{eq:promote}
  U_n(\cand_k) - \E_\tau[X(\tau)] \le s_k
  \;\Longleftrightarrow\;
  K_n(m_\ell) \ge \frac{1}{\alpha_k} .
\end{equation}
\begin{proposition}[The data cost of a test]
\label[proposition]{cor:rate}
Let $\sigma_k^2 = \mathrm{Var}_{\tau \sim (\pi_k, \cM)}[X(\tau)]$ be the variance of the statistic at test $k$. Let $\Gamma = w_X - \ell + c_{>H}$ be the \emph{headroom}, $\lambda_{\max} = 1/(2(1 - m_\ell))$ the bet cap of \cref{app:betting}, and $\kappa = 1 + \lambda_{\max} m_\ell = (w_X + \Gamma)/(2\Gamma)$. Suppose $\Gamma > 0$, condition on everything observed before the stream of test $k$ starts and on the simulation randomness of that stream, and let the candidate have slack $s_k > 0$. Promotion time is then lower bounded:
\begin{equation}
  \label{eq:sample-complexity}
  \E[N_P] \;\ge\; 2\ln\frac{1}{\alpha_k} \max\left(\frac{1}{\kappa}\left(1 + \frac{\sigma_k^2}{s_k^2}\right),\, \frac{\Gamma}{s_k}\right),
\end{equation}
  and the fixed bet $\lambda^\circ = \frac{w_X}{2}\min\left(\frac{s_k}{\sigma_k^2 + s_k^2}, \frac{1}{\Gamma}\right)$, placed at every trajectory, gives a promotion time upper bound:
\begin{equation}
  \label{eq:cost-upper}
  \E[N_P] \;\le\; 4\left(\ln\frac{1}{\alpha_k} + \ln\kappa\right) \max\left(1 + \frac{\sigma_k^2}{s_k^2},\, \frac{\Gamma}{s_k}\right).
\end{equation}
The data cost satisfies $N_k \le N_P$. See \cref{app:rate} for a proof.
\end{proposition}

The bounds of \cref{cor:rate} have tighter forms in terms of the growth rate of the capital (\cref{lem:passage}); we weaken them into functions of the slack $s_k$, variance $\sigma_k^2$ and headroom $\Gamma$ to show which of these set the data cost. The \emph{range regime} ($\Gamma/s_k$) is set by the headroom. The \emph{variance regime} ($\sigma_k^2/s_k^2$) has two sources: the spread of the clipped cumulative importance weights $\cWc(\tau)$ and the noise of the costs they multiply. The anchor of \cref{eq:proposer} reduces the first: at each state the per-step log-weight has variance $\mathrm{Var}_{a \sim \pi_k(\cdot \mid s)}[\ln \rho(s, a)] \approx 2\, D_{\mathrm{KL}}(\cand_k \,\|\, \pi_k)$ (\cref{app:perstep}), so a small anchor value keeps the cumulative importance weight concentrated near one. A doubly-robust statistic \citep{jiang2016doubly, thomas2016data} could trade the second for model noise; \cref{app:dr} records why Dejavu does not use one.

A test with promotion time in \cref{eq:stopping} that satisfies $N_P > \Nmax$ wastes $\Nmax$ production trajectories. Some such candidates are unsafe, but many are safe with too little slack to certify within the budget. Given the test passes on \cref{eq:promote}, we determine whether $\ln K_n(m_\ell)$ can reach $\ln(1/\alpha_k)$ in the $\Nmax - n$ trajectories that are left. Let the \emph{growth rate} $g(\lambda) = \E[\ln(1 + \lambda(m_\ell - Y))]$ be the expected increase of $\ln K_n(m_\ell)$ per trajectory at bet $\lambda$. By \cref{app:rate}, a promotion needs $\E[N_P] \ge \ln(1/\alpha_k)/g^\star$ trajectories, with $g^\star = \max_{\lambda \in [0, \lambda_{\max}]} g(\lambda)$. The test cannot observe $g^\star$, but it can bound it. Let $\hat g_n$ be an anytime-valid upper confidence betting bound on $g^\star$ at level $\beta$, so $\Prob(\exists n : g^\star > \hat g_n) \le \beta$. The test refuses a candidate at the first buffer size $n$ with the \emph{early refusal rule}
\begin{equation}
  \label{eq:early-refusal}
  \ln K_n(m_\ell) + (\Nmax - n)\, \hat g_n \;<\; \ln(1/\alpha_k),
\end{equation}
The guarantee this buys is \cref{prop:early}.
\begin{proposition}[Early refusal]
\label[proposition]{prop:early}
Given the early refusal rule of \cref{eq:early-refusal}, for every $n$ and every $N$ with $n \le N \le \Nmax$,
\begin{equation}
  \label{eq:expected-capital}
  \E\big[\ln K_N(m_\ell) \,\big|\, \cF_n\big] \;\le\; \ln K_n(m_\ell) + (N - n)\, \hat g_n ,
\end{equation}
so a test has expected log-capital below $\ln(1/\alpha_k)$ at every buffer size it can reach. See \cref{app:early} for a proof.
\end{proposition}

\cref{eq:expected-capital} is an expectation statement, so a candidate that would have passed by a favorable run of trajectories can be refused. The cost of that error is a lost candidate rather than safety, so $\beta$ is accepted as a per-test rate, not a lifetime budget.

The production and simulation environments differ only in the transition kernel and the initial state distribution (\cref{sec:prelim}), so we define simulation quality by the distance between them. For a candidate $\cand$ in the simulation environment, let $\tilde d_{\cand}(s, a) = (1 - \gc) \sum_{t \ge 0} \gc^t\, \Prob_{\tau \sim (\cand, \tilde\cM)}(s_t = s, a_t = a)$ be its \emph{discounted state-action visitation}, and let the \emph{total variation distance} be denoted $D_\mathrm{TV}(p, q) = \tfrac{1}{2}\|p - q\|_1$. The \emph{initial-state gap} and the \emph{dynamics gap} of $\cand$ are $\varepsilon_0 = D_\mathrm{TV}(d_0, \tilde d_0)$ and $\varepsilon_T(\cand) = \E_{(s, a) \sim \tilde d_{\cand}}[D_\mathrm{TV}\big( T(\cdot \mid s, a),\, \simT(\cdot \mid s, a) \big)]$ respectively, and their combination is the \emph{sim-to-real gap} \citep{kearns2002near, agarwal2019reinforcement}
\begin{equation}
  \label{eq:gap}
  \varepsilon(\cand) = \frac{\cmax}{1 - \gc} \left( \varepsilon_0 + \frac{\gc\, \varepsilon_T(\cand)}{1 - \gc} \right).
\end{equation}
A candidate returned by the proposer of OP (\ref{eq:proposer}) satisfies the simulation constraint, $\tilde J_C(\cand_k) \le \tilde \ell$, and the \emph{simulation margin} is given by $\ell - \tilde \ell$.

\begin{proposition}[Slack under the sim-to-real gap]
\label[proposition]{prop:gap}
Under \cref{ass:bounded,ass:cap}, a candidate that satisfies the simulation constraint, $\tilde J_C(\cand_k) \le \tilde \ell$, has slack
\begin{equation}
  \label{eq:slack-bound}
  \ell - J_C(\cand_k) \ge s_k \;\ge\; (\ell - \tilde \ell) - c_{>H} - \varepsilon(\cand_k).
\end{equation}
See \cref{app:gap} for a proof.
\end{proposition}

The upper bound says an unsafe candidate has negative slack and cannot be certified safe with any amount of data. The lower bound shows a larger sim-to-real gap or a smaller simulation margin lowers the slack a candidate is guaranteed, which raises the bound on its test's data cost in \cref{cor:rate}.

\begin{figure*}[t]
\centering
\includegraphics[width=\textwidth]{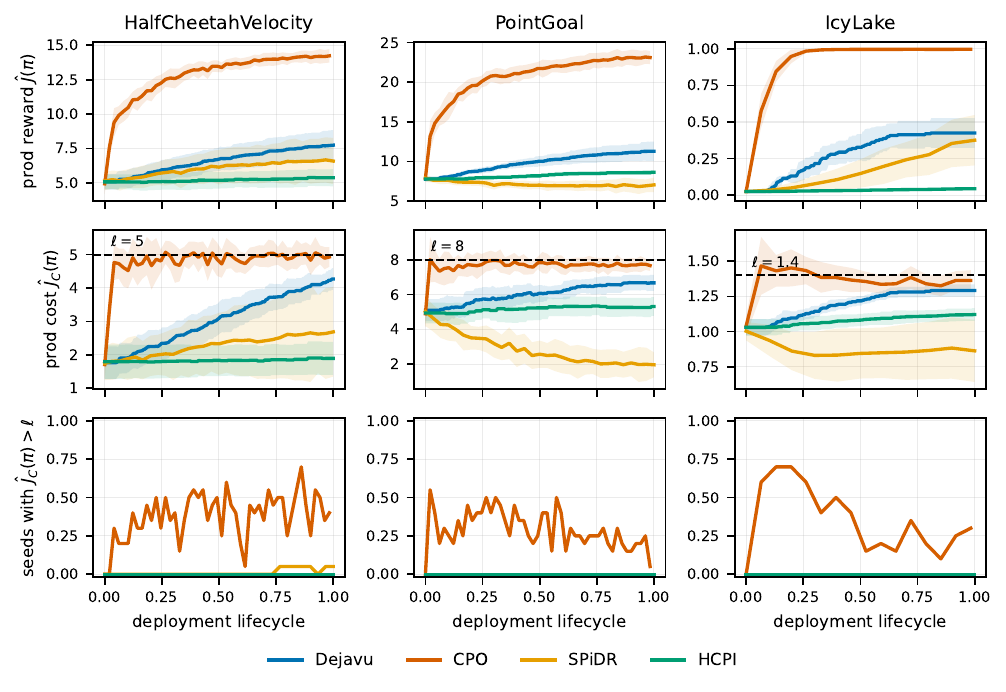}
\caption{Deployment benchmark, one column per environment: the baseline policy's true production-environment reward (top) and cost return with the limit (middle), and the share of seeds whose current baseline policy exceeds the cost constraint (bottom). Lines are means over $20$ seeds, shaded regions are $\pm1$ standard deviation. The normalized deployment lifecycle describes the fraction of the total production data budget consumed for training by each method. SPiDR uses 0 production data; the others have a budget of $10$M steps on the CRAX environments and $1$M steps on IcyLake.}
\label{fig:benchmark}
\end{figure*}

\section{Experiments}
\label{sec:experiments}
We evaluate on three sim-to-real CMDPs, each a JAX implementation of a recognized benchmark so that training runs on the GPU, which lets us report results over $20$ training runs initialized with different random number generator seeds. Two come from CRAX \citep{tomilin2026crax}, a JAX library of constrained locomotion and navigation tasks built on the MuJoCo XLA (MJX) physics engine.
\begin{itemize}[nosep, itemsep=2pt]
    \item \textbf{HalfCheetahVelocity}: Safety Gymnasium's \citep{ji2023safety} SafetyHalfCheetahVelocity, as implemented in CRAX. The agent maximizes forward velocity, and incurs cost $1$ when its forward speed exceeds $3.21$\,m/s.
    \item \textbf{PointGoal}: SafetyPointGoal1 from Safety Gymnasium \citep{ray2019benchmarking, ji2023safety}, as implemented in CRAX. The agent navigates to a randomized goal location, and incurs cost $1$ when it enters one of $16$ hazard regions.
    \item \textbf{IcyLake}: A JAX reimplementation of the Gymnasium FrozenLake environment \citep{towers2024gymnasium}. The agent must cross a $4 \times 4$ frozen lake grid to the goal state while avoiding icy tiles, each of which costs $1$ per visit.
\end{itemize}
Each pairs a production environment with a deliberately optimistic simulation environment. Each environment runs at one cost limit $\ell$ chosen so that the constraint binds. All runs use the budget of \cref{eq:budget} with $\alpha = 0.075$ and $\alpha_{\pi_0} = 0.025$, and the initial policy is certified on-policy (\cref{ass:start}). Environment parameters, cost limits, and hyperparameters are given in \cref{app:experiments}.

We compare four ways of turning the same safe initial policy into a sequence of baseline policies.
\begin{itemize}[nosep, itemsep=2pt]
  \item \textbf{CPO} \citep{achiam2017constrained} updates the policy directly in the production environment with a worst-case safety constraint violation magnitude at each update.
  \item \textbf{SPiDR} \citep{as2025spidr} each round retrains from the current baseline policy with CRPO \citep{xu2021crpo} in simulation under domain randomization with a pessimism penalty, and deploys every retrained policy without a test.
  \item \textbf{Dejavu} proposes each candidate with CRPO \citep{xu2021crpo} in the simulation environment and promotes it only if it passes the test of \cref{sec:certificate}.
  \item \textbf{HCPI} \citep{thomas2015improvement} each iteration collects production trajectories of the baseline policy, on which a fifth are used to propose a candidate via CPO. The remainder are used for the test, with the high-confidence upper bound of \citet{thomas2015high}.
\end{itemize}
Ground truth for every deployed policy is a held-out Monte-Carlo estimate of its expected production cost return, denoted $\hat J_C(\pi)$ (\cref{app:experiments}).

Dejavu satisfies its lifetime validity constraint for every environment in \cref{fig:benchmark}, while accumulating more reward than the other safe methods. Its closest counterpart, HCPI, is also safe, but barely moves at all. This is expected given the current literature highlights sample efficiency and computational complexity as its main drawbacks \citep{thomas2015improvement}. CPO trained in production reaches the highest reward levels, but frequently deploys policies that violate the cost limit. This is expected since CPO only bounds the worst-case constraint violation magnitude at each update, so in theory every policy could violate the constraint. The only other method that trains candidates in simulation is SPiDR, which makes it a useful comparison to Dejavu on reward accumulation. SPiDR also stays safe across environments, but has a lower reward than Dejavu. It should also be noted that to satisfy safety, SPiDR uses a pessimism penalty that is tuned for each production environment, while Dejavu is agnostic to simulation quality and did not require the same tuning (\cref{app:experiments}).

\section{Conclusion}
\label{sec:conclusion}

Dejavu turns the sim-to-real deployment decision into a test: a candidate policy trained in simulation is promoted only when an anytime-valid upper confidence bound on its expected cost return in the production environment is below the cost limit. The test allows for an adaptive number of trajectories to be sampled by a baseline policy in the production environment to reach promotion, and a summable budget extends the guarantee over an infinite promotion sequence. The method is agnostic to how candidates are produced, and simulation quality only affects sample efficiency, not the validity of the deployment decision. Finally, experimental results support the theoretical guarantees, where Dejavu promotes safe candidates with superior reward accumulation compared to other safe methods.

\section{Limitations and Future Work}
\label{sec:future}

Dejavu is a first step towards systems that improve a deployed policy continuously while keeping a statistically valid safety guarantee over the whole deployment lifecycle. Its main limitation is sample efficiency, and determining how far it can be improved is the subject of ongoing work. Dejavu also assumes independent trajectories from a fixed initial state distribution under fixed production dynamics. Many production systems under continuous operation instead operate without resets, and their dynamics change over time; extending lifetime validity to these systems is a further direction.

\section*{Acknowledgements}

Rohan Tangri gratefully acknowledges support from G-Research.

\bibliography{dejavu}
\bibliographystyle{icml2026}

\newpage
\appendix
\onecolumn

\section{Proofs}
\label[appendix]{app:proofs}

\subsection{Proof of \cref{lem:envelope}}
\label[appendix]{app:envelope}

Throughout, we abbreviate $\pi = \pi_{k}$, $\cand = \cand_k$. \cref{ass:cap} implies in particular that $\pi(a \mid s) > 0$ wherever $\cand(a \mid s) > 0$ along trajectories of $(\pi, \cM)$, so all per-step importance weights below are well defined.

Write the candidate's cost return and split at the horizon:
\begin{equation}
  J_C(\cand) = \E_{\tau \sim (\cand, \cM)}\Big[\sum_{t=0}^{H-1} \gc^t c_t\Big] + \E_{\tau \sim (\cand, \cM)}\Big[\sum_{t=H}^{\infty} \gc^t c_t\Big] .
\end{equation}
For the head, the PDIS identity \citep{precup2000eligibility} gives
\begin{equation}
  \E_{\tau \sim (\cand, \cM)}\Big[\sum_{t=0}^{H-1} \gc^t c_t\Big] = \E_{\tau \sim (\pi, \cM)}\Big[\sum_{t=0}^{H-1} \gc^t\, \cW(\tau)\, c_t\Big] .
\end{equation}
Under \cref{ass:cap} the clip almost surely never binds, so $\cWc = \cW$ almost surely, and the right-hand side equals $\E[X(\tau)]$. For the tail, no importance sampling is involved. It is an expectation under $(\cand, \cM)$ directly, and \cref{ass:bounded} gives
\begin{equation}
  \E_{\tau \sim (\cand, \cM)}\Big[\sum_{t=H}^{\infty} \gc^t c_t\Big]
  \le \sum_{t=H}^{\infty} \gc^t \cmax
  = \frac{\gc^H \cmax}{1 - \gc} .
\end{equation}
The last expression is the truncation offset $c_{>H}$ defined after \cref{eq:certvalue}, so combining the two displays yields $J_C(\cand) \le \E[X(\tau)] + c_{>H}$.

Each term of \cref{eq:certvalue} is a product of three nonnegative factors bounded by $\gc^t$, $\Wmax$, and $\cmax$ respectively, so
\begin{equation}
  0 \le X(\tau) \le \Wmax \cmax \sum_{t=0}^{H-1} \gc^t
  = \Wmax \cmax \frac{1 - \gc^H}{1 - \gc} = w_X .
\end{equation}

\subsection{Proof of \cref{prop:clip}}
\label[appendix]{app:clipproof}

Throughout, $\pi = \pi_k$ is the baseline policy, $\cand = \cand_k$ the candidate, $c_t \in [0, \cmax]$ by \cref{ass:bounded}, and expectations are over $\tau \sim (\pi, \cM)$ unless marked otherwise. We assume $\pi(a \mid s) > 0$ wherever $\cand(a \mid s) > 0$, so every per-step importance weight $\rho(s, a) = \cand(a \mid s) / \pi(a \mid s)$ of \cref{eq:per-step-is} is finite.

The statistic $X(\tau)$ of \cref{eq:certvalue} uses the clipped cumulative importance weight $\cWc(\tau) = \min(\cW(\tau), \Wmax)$ instead of $\cW(\tau)$. The \emph{clip bias} is the gap between the truncated cost return of the candidate and the mean of the statistic:
\begin{equation}
\label{eq:cum-shortfall}
  \begin{aligned}
    J^H_C(\cand) - \E[X(\tau)]
    &= \E_{\tau \sim (\cand, \cM)}\Big[\sum_{t<H} \gc^t c_t\Big] - \E\Big[\sum_{t<H} \gc^t\, \cWc(\tau)\, c_t\Big]
    \\
    &= \E\Big[\sum_{t<H} \gc^t\, \cW(\tau)\, c_t\Big] - \E\Big[\sum_{t<H} \gc^t\, \cWc(\tau)\, c_t\Big]
    && \text{[PDIS, \cref{eq:pdis}]} \\
    &= \sum_{t<H} \gc^t\, \E\big[(\cW(\tau) - \cWc(\tau))\, c_t\big]
    \\
    &= \sum_{t<H} \gc^t\, \E\big[(\cW(\tau) - \Wmax)^+\, c_t\big].
  \end{aligned}
\end{equation}
Every term of the last line of \cref{eq:cum-shortfall} is nonnegative, because $(\cW(\tau) - \Wmax)^+ \ge 0$ and $c_t \ge 0$, so $J^H_C(\cand) - \E[X(\tau)] \ge 0$.

The per-step importance weight has conditional mean one: at any state $s$,
\begin{equation}
\label{eq:unit-mean}
  \E_{a \sim \pi(\cdot \mid s)}[\rho(s, a)] = \sum_a \pi(a \mid s)\, \frac{\cand(a \mid s)}{\pi(a \mid s)} = \sum_a \cand(a \mid s) = 1 ,
\end{equation}
with the sum replaced by an integral for continuous actions. Applying \cref{eq:unit-mean} to the last factor of $\cW(\tau) = \prod_{i \le t} \rho(s_i, a_i)$, conditional on the trajectory up to $s_t$, and repeating for each earlier factor gives
\begin{equation}
\label{eq:unit-mass}
  \E[\cW(\tau)] = 1 \quad \text{for every } t .
\end{equation}
Then
\begin{equation}
  \begin{aligned}
    J^H_C(\cand) - \E[X(\tau)]
    &= \sum_{t<H} \gc^t\, \E\big[(\cW(\tau) - \cWc(\tau))\, c_t\big]
    && \text{[\cref{eq:cum-shortfall}]} \\
    &\le \cmax \sum_{t<H} \gc^t\, \E\big[\cW(\tau) - \cWc(\tau)\big]
    && \text{[$c_t \le \cmax$, $\cW(\tau) - \cWc(\tau) \ge 0$]} \\
    &= \cmax \sum_{t<H} \gc^t\, (1 - \varrho_t)
    && \text{[\cref{eq:unit-mass}, $\varrho_t = \E[\cWc(\tau)]$]} .
  \end{aligned}
\end{equation}
Under \cref{ass:cap}, $\cW(\tau) \le \Wmax$ almost surely, so $\cWc = \cW$, $\varrho_t = 1$ by \cref{eq:unit-mass}, and the clip bias is zero.

\subsection{Proof of \cref{thm:rung}}
\label[appendix]{app:rung}

Fix test $k$ and condition on $\cF_{<k}$ throughout; every probability below is conditional on it.

By \cref{ass:indep}, $\cand_k$ is fixed given $\cF_{<k}$ and is never trained on $D_k$, so the buffer trajectories $\tau_1, \tau_2, \ldots$ are drawn i.i.d.\ from $(\pi_k, \cM)$ with the candidate held fixed. The values of the statistic, $X(\tau_1), X(\tau_2), \ldots$ are therefore i.i.d.\ with mean
\begin{equation}
  \mu_k \coloneqq \E_{\tau \sim (\pi_k, \cM)}\big[X(\tau) \,\big|\, \cF_{<k}\big],
\end{equation}
and by \cref{eq:certvalue}, which needs only \cref{ass:bounded} since the clip caps the weight at $\Wmax$, they lie in $[0, w_X]$. Hence $Y_j = X(\tau_j)/w_X$ are i.i.d.\ in $[0, 1]$ with mean $\mu_k / w_X$, which is the setting of the betting confidence sequence in \cref{sec:prelim}.

Run the betting confidence sequence of \cref{eq:capital,eq:ucs} on $Y_1, Y_2, \ldots$ at level $\alpha_k$. Its guarantee, $\Prob(\exists n : \mu_k / w_X > u_n) \le \alpha_k$, holds for any predictable bet schedule with $\lambda_j(m) \in [0, \tfrac{1}{1-m}]$ \citep{ville1939etude, waudby2024estimating}. Multiplying through by $w_X$ and using $U_n(\cand_k) = w_X\, u_n$ from \cref{eq:Un},
\begin{equation}
  \label{eq:proof-anytime}
  \Prob\big( \exists n : \mu_k > U_n(\cand_k) \,\big|\, \cF_{<k} \big) \le \alpha_k .
\end{equation}

Writing $\{\cdot\}$ for the event that the enclosed inequality holds, the event in \cref{eq:proof-anytime} is a union over all $n$, so it contains the same event at any data-dependent index, in particular at the stopping time $N_k$ of \cref{eq:stopping}:
\begin{equation}
  \big\{ \mu_k > U_{N_k}(\cand_k) \big\} \subseteq \big\{ \exists n : \mu_k > U_n(\cand_k) \big\}.
\end{equation}
This is the only place the anytime-valid property is used, and it is why the test may stop by any rule without affecting the guarantee. Finally, \cref{lem:envelope} gives $J_C(\cand_k) \le \mu_k + c_{>H}$ for the fixed candidate, so the bound failure event has property:
\begin{equation}
  \big\{ J_C(\cand_k) > U_{N_k}(\cand_k) + c_{>H} \big\} \subseteq \big\{ \mu_k > U_{N_k}(\cand_k) \big\}.
\end{equation}
Chaining the two inclusions and applying \cref{eq:proof-anytime},
\begin{equation}
  \Prob\big( J_C(\cand_k) > U_{N_k}(\cand_k) + c_{>H} \,\big|\, \cF_{<k} \big)
  \le \Prob\big( \exists n : \mu_k > U_n(\cand_k) \,\big|\, \cF_{<k} \big)
  \le \alpha_k .
\end{equation}

\subsection{Proof of \cref{thm:lifetime}}
\label[appendix]{app:lifetimeproof}

Group the tests into streams: stream $i$ consists of the tests $k$ with $i_k = i$, all of which share one baseline policy $\pi^{(i)}$ and one buffer $D^{(i)}$, with the stream ending at its first promotion. Write $\cand_{i,j}$ and $\alpha_{i,j}$ for the candidate and the level of the test at position $p_k = j$ of stream $i$, $U^{(i,j)}_n$ for the bound of that test on the first $n$ trajectories of $D^{(i)}$, and $j_i$ for the position at which stream $i$ promotes. A \emph{run} is one execution of \cref{alg:dejavu} with any such test; the events below are sets of runs. Let $\cF^{(i)}$ contain all data, candidates, and decisions before the first test of stream $i$, and the randomness of the simulation environment during stream $i$. The candidates $\cand_{i,j}$ depend only on $\cF^{(i)}$ by \cref{ass:indep}, as do the levels $\alpha_{i,j}$ by assumption (checked for Dejavu at the end of the proof). Furthermore, trajectories of $D^{(i)}$ are i.i.d.\ samples of $(\pi^{(i)}, \cM)$ given $\cF^{(i)}$. So by assumption the test satisfies
\begin{equation}
  \label{eq:stream-anytime}
  \Prob\big( \exists n : J_C(\cand_{i,j}) > U^{(i,j)}_n + c_{>H} \,\big|\, \cF^{(i)} \big) \le \alpha_{i,j} .
\end{equation}

Every baseline policy is either $\pi_0$ or a candidate $\cand_{i,j_i}$ that ended some stream $i$ by promotion. Hence
\begin{equation}
  \label{eq:lifetime-split}
  \Prob\big( \exists\, k : J_C(\pi_k) > \ell \big)
  = \Prob\Big( \{ J_C(\pi_0) > \ell \} \cup \bigcup_{i \ge 0} F_i \Big)
  \le \alpha_{\pi_0} + \Prob\Big( \bigcup_{i \ge 0} F_i \Big),
\end{equation}
where $F_i \coloneqq \{ \text{stream } i \text{ promotes a candidate with } J_C(\cand_{i,j_i}) > \ell \}$ is a false promotion in stream $i$, and the inequality is a union bound with \cref{ass:start}. It remains to show that the probability of a run containing any false promotions is bounded: $\Prob(\bigcup_i F_i) \le \alpha$.

Let $Q_i$ be the event that no false promotion occurred in streams $0, \ldots, i-1$. The event $Q_i$ is decided by the streams before $i$, so it is $\cF^{(i)}$-measurable. Then
\begin{equation}
  \label{eq:lifetime-first}
  \begin{aligned}
    \Prob\Big( \bigcup_{i \ge 0} F_i \Big)
    &= \Prob\Big( \bigcup_{i \ge 0} (F_i \cap Q_i) \Big)
    && \text{[a run with a false promotion has a first one]} \\
    &= \sum_{i \ge 0} \Prob(F_i \cap Q_i)
    && \text{[$F_i \cap Q_{i'} = \emptyset$ for $i < i'$]} \\
    &= \sum_{i \ge 0} \E\big[ \mathbf{1}_{Q_i} \mathbf{1}_{F_i} \big] \\
    &= \sum_{i \ge 0} \E\Big[ \E\big[ \mathbf{1}_{Q_i} \mathbf{1}_{F_i} \,\big|\, \cF^{(i)} \big] \Big]
    && \text{[tower property]} \\
    &= \sum_{i \ge 0} \E\big[ \mathbf{1}_{Q_i}\, \Prob(F_i \mid \cF^{(i)}) \big]
    && \text{[$Q_i$ is $\cF^{(i)}$-measurable]} .
  \end{aligned}
\end{equation}

Fix $i$. Let the false promotion event at position $j$ of stream $i$ be
\begin{equation}
  \label{eq:false-promotion}
  E_{i,j} = \big\{ j_i = j \big\} \cap \big\{ J_C(\cand_{i,j}) > \ell \big\},
\end{equation}
so that $F_i = \bigcup_{j \ge 0} E_{i,j}$. A false promotion at position $j$ has three necessary conditions that are jointly sufficient: stream $i$ reaches position $j$, so no earlier candidate in the stream was promoted; the test at position $j$ passes, so $U^{(i,j)}_n + c_{>H} \le \ell$ at a buffer size $n$ that the test checks before it refuses or reaches its data budget; and the candidate is unsafe, so $J_C(\cand_{i,j}) > \ell$. Dropping the first condition, and keeping from the second only that $U^{(i,j)}_n + c_{>H} \le \ell$ at some $n$, leaves two conditions that are necessary but not sufficient. Together they give $U^{(i,j)}_n + c_{>H} \le \ell < J_C(\cand_{i,j})$, so the bound fails at that $n$, and
\begin{equation}
  \label{eq:false-promotion-subset}
  E_{i,j} \subseteq \big\{ \exists n : J_C(\cand_{i,j}) > U^{(i,j)}_n + c_{>H} \big\} \cap \big\{ J_C(\cand_{i,j}) > \ell \big\} .
\end{equation}
The event on the right is larger than $E_{i,j}$: it also contains $\big\{ \exists n : J_C(\cand_{i,j}) > U^{(i,j)}_n + c_{>H} \big\} \cap \big\{ J_C(\cand_{i,j}) > \ell \big\} \cap \big\{ j_i \ne j \big\}$,
runs in which the bound of an unsafe candidate fails but the candidate is refused or never tested. Then
\begin{equation}
  \label{eq:stream-bound}
  \begin{aligned}
    \Prob(F_i \mid \cF^{(i)})
    &= \Prob\Big( \bigcup_{j \ge 0} E_{i,j} \,\Big|\, \cF^{(i)} \Big)
    && \text{[\cref{eq:false-promotion}]} \\
    &\le \Prob\Big( \bigcup_{j \ge 0} \Big( \big\{ \exists n : J_C(\cand_{i,j}) > U^{(i,j)}_n + c_{>H} \big\} \cap \big\{ J_C(\cand_{i,j}) > \ell \big\} \Big) \,\Big|\, \cF^{(i)} \Big)
    && \text{[\cref{eq:false-promotion-subset}]} \\
    &\le \sum_{j \ge 0} \Prob\Big( \big\{ \exists n : J_C(\cand_{i,j}) > U^{(i,j)}_n + c_{>H} \big\} \cap \big\{ J_C(\cand_{i,j}) > \ell \big\} \,\Big|\, \cF^{(i)} \Big)
    && \text{[union bound]} \\
    &= \sum_{j \ge 0} \mathbf{1}\{ J_C(\cand_{i,j}) > \ell \}\, \Prob\big( \exists n : J_C(\cand_{i,j}) > U^{(i,j)}_n + c_{>H} \,\big|\, \cF^{(i)} \big)
    && \text{[$\cand_{i,j}$ fixed given $\cF^{(i)}$]} \\
    &\le \sum_{j \ge 0} \alpha_{i,j}\, \mathbf{1}\{ J_C(\cand_{i,j}) > \ell \}
    && \text{[\cref{eq:stream-anytime}]} .
  \end{aligned}
\end{equation}
The union runs over every position the stream could reach, reached or not, because whether position $j$ is reached depends on $D^{(i)}$. This is why the levels within a stream must be fixed when it opens and why the levels of tests that never happen are charged. Substituting \cref{eq:stream-bound} into \cref{eq:lifetime-first}, and exchanging the sum and the expectation by monotone convergence,
\begin{equation}
  \label{eq:lifetime-expect}
  \Prob\Big( \bigcup_{i \ge 0} F_i \Big) \;\le\; \E\Big[ \sum_{i \ge 0} \mathbf{1}_{Q_i} \sum_{j \ge 0} \alpha_{i,j}\, \mathbf{1}\{ J_C(\cand_{i,j}) > \ell \} \Big].
\end{equation}

Fix one run and let $i^\star$ be its first stream with a false promotion, $i^\star = \infty$ if there is none, so that $\mathbf{1}_{Q_i} = 1$ for $i \le i^\star$ and $\mathbf{1}_{Q_i} = 0$ for $i > i^\star$. Let $A_i = \sum_{j \ge 0} \alpha_{i,j}$ be the \emph{stream budget}, the sum of the levels at every position of stream $i$, reached or not. For $i < i^\star$, the promotion that ended stream $i$ was safe, so $\mathbf{1}\{ J_C(\cand_{i,j_i}) > \ell \} = 0$, and every other indicator is at most $1$. Then
\begin{equation}
  \label{eq:lifetime-run}
  \begin{aligned}
    &\sum_{i \ge 0} \mathbf{1}_{Q_i} \sum_{j \ge 0} \alpha_{i,j}\, \mathbf{1}\{ J_C(\cand_{i,j}) > \ell \} \\
    &\quad = \sum_{i \le i^\star} \sum_{j \ge 0} \alpha_{i,j}\, \mathbf{1}\{ J_C(\cand_{i,j}) > \ell \}
    && \text{[definition of $i^\star$]} \\
    &\quad = \sum_{i < i^\star} \sum_{j \ge 0} \alpha_{i,j}\, \mathbf{1}\{ J_C(\cand_{i,j}) > \ell \}
     + \sum_{j \ge 0} \alpha_{i^\star,j}\, \mathbf{1}\{ J_C(\cand_{i^\star,j}) > \ell \}
    && \text{[split off $i = i^\star$]} \\
    &\quad \le \sum_{i < i^\star} \sum_{j \ne j_i} \alpha_{i,j} + \sum_{j \ge 0} \alpha_{i^\star,j}
    && \text{[safe promotions before $i^\star$]} \\
    &\quad = \sum_{i < i^\star} \big( A_i - \alpha_{i, j_i} \big) + A_{i^\star}
    && \text{[definition of $A_i$]} .
  \end{aligned}
\end{equation}
Let $I_i$ be the injections assigned to stream $i$, as defined in \cref{sec:ladder}. Assume $\sum_i I_i \le \alpha$ and that the stream budget follows the \emph{fallback rule}
\begin{equation}
  \label{eq:fallback}
  A_0 = I_0, \qquad A_{i+1} = I_{i+1} + \alpha_{i, j_i} ,
\end{equation}
that is, a stream's levels sum to its own injection plus the level of the test that promoted into it. Then $A_i - \alpha_{i,j_i} = A_i - A_{i+1} + I_{i+1}$, and for $i^\star < \infty$ the sum telescopes:
\begin{equation}
  \sum_{i < i^\star} \big( A_i - \alpha_{i, j_i} \big) + A_{i^\star}
  = A_0 - A_{i^\star} + \sum_{i = 1}^{i^\star} I_i + A_{i^\star}
  = \sum_{i \le i^\star} I_i \;\le\; \alpha .
\end{equation}
If $i^\star = \infty$, the partial sum up to any $m$ is $A_0 - A_m + \sum_{i=1}^{m} I_i \le \sum_{i \le m} I_i \le \alpha$, since $A_m \ge 0$, so the same bound holds.

This bound holds on every run, so \cref{eq:lifetime-expect} gives $\Prob(\bigcup_i F_i) \le \alpha$, and \cref{eq:lifetime-split} gives $\Prob(\exists\, k : J_C(\pi_k) > \ell) \le \alpha_{\pi_0} + \alpha$.

It remains to check that Dejavu meets these conditions. Its test satisfies \cref{eq:stream-anytime} under \cref{ass:bounded,ass:cap}, by \cref{eq:proof-anytime} with \cref{lem:envelope}. \cref{alg:dejavu} has the three properties of the level schedule assumed above. Its injections $I_i = \eta_i\alpha$ sum to $\alpha$ by \cref{eq:budget}. Its levels, set by \cref{eq:levels}, in stream $i \ge 1$ sum to
\begin{equation}
  \alpha_{i-1,j_{i-1}} + (1 - \phi_i)\,\eta_i\alpha + \phi_i\,\eta_i\alpha \sum_{p \ge 1} \frac{1}{p(p+1)} = I_i + \alpha_{i-1,j_{i-1}},
\end{equation}
which is the fallback rule \cref{eq:fallback}, and those of stream $0$ sum to $I_0$ since $\alpha_{-1} = 0$. The levels \cref{eq:levels} of stream $i$ depend only on the constant $\eta_i$, on the passed level $\alpha_{i-1,j_{i-1}}$, which is set when stream $i-1$ promotes, and on the reserve fraction $\phi_i$ of \cref{eq:reserve}, which depends only on $i$ and on the index $k$ of the stream's first test. All three are determined before the first trajectory of $D^{(i)}$ is collected, so the levels are $\cF^{(i)}$-measurable.

\subsection{Proof of \cref{cor:rate}}
\label[appendix]{app:rate}

Let test $k$ sit at position $j$ of stream $i$, and condition on $\cF^{(i)}$, everything observed before the stream starts together with the randomness of the simulation environment during the stream (\cref{app:lifetimeproof}). The candidate $\cand_k$ is then fixed by \cref{ass:indep}, the level $\alpha_k$ is fixed when the stream starts (\cref{eq:levels}), and the values $Y_j = X(\tau_j)/w_X \in [0, 1]$ are i.i.d, since the trajectories of the stream are i.i.d.\ samples of $(\pi_k, \cM)$. The promotion time $N_P$ of \cref{eq:stopping} is the number of production trajectories required to promote the candidate, and we wish to obtain upper and lower bounds on it. 

The test works in units of $Y$ (\cref{eq:capital}), so each quantity of \cref{cor:rate} has a normalized counterpart, obtained by dividing by the range $w_X$ of the statistic:
\begin{equation}
  \label{eq:normalized}
  \begin{aligned}
    m_\ell &= \frac{\ell - c_{>H}}{w_X}, &\qquad \bar s &= \frac{s_k}{w_X} = m_\ell - \E[Y], \\
    1 - m_\ell &= \frac{\Gamma}{w_X}, &\qquad \bar q &= \E\big[(m_\ell - Y)^2\big] = \frac{\sigma_k^2 + s_k^2}{w_X^2},
  \end{aligned}
\end{equation}
the normalized threshold of \cref{eq:promote}, and the \emph{normalized slack}, \emph{normalized headroom} and \emph{normalized second moment} of $Y$ about $m_\ell$. The mean, slack and headroom split the range, $\E[Y] + \bar s + (1 - m_\ell) = 1$, and $m_\ell \ge \bar s > 0$ because $\E[Y] \ge 0$. In the capital of \cref{eq:capital} at $m = m_\ell$, each trajectory gains $\lambda(m_\ell - Y)$ at bet $\lambda$, to first order: the gain per unit bet has mean $\bar s$ and second moment $\bar q$, and its largest loss, at $Y = 1$, is the headroom $1 - m_\ell$, which sets the bet cap $\lambda_{\max} = 1/(2(1 - m_\ell))$. The ratios of these quantities that enter \cref{cor:rate} do not depend on $w_X$:
\begin{equation}
  \label{eq:rate-units}
  \frac{\bar q}{\bar s^2} = 1 + \frac{\sigma_k^2}{s_k^2}, \qquad \lambda_{\max} \bar s = \frac{s_k}{2\Gamma}, \qquad 1 + \lambda_{\max} m_\ell = \kappa .
\end{equation}
The \emph{growth rate} $g(\lambda) = \E[\ln(1 + \lambda(m_\ell - Y))]$ is the expected increase of the log-capital per trajectory at bet $\lambda$. The test's bets are $\lambda_j = \lambda_j(m_\ell) \in [0, \lambda_{\max}]$, since $\lambda_{\max}$ is the cap of \cref{app:betting} at $m = m_\ell$. The test promotes at the first $n$ with $\ln K_n(m_\ell) \ge \ln(1/\alpha_k)$. For $\lambda \in [0, \lambda_{\max}]$ the argument $x = \lambda(m_\ell - Y)$ of the growth rate lies in $[-1/2,\, \lambda m_\ell]$, since $Y \in [0, 1]$ and $\lambda_{\max}(1 - m_\ell) = 1/2$.

\begin{lemma}[Expected data at a given growth rate]
\label[lemma]{lem:passage}
Condition on the start of the candidate's stream, so that the candidate is fixed and the values of the statistic are i.i.d.\ (as above). The promotion time $N_P$ of \cref{eq:stopping} is the first buffer size at which the test condition holds, so by \cref{eq:promote} it equals $\min\{n : \ln K_n(m_\ell) \ge \ln(1/\alpha_k)\}$. 

(a) For any predictable bet schedule with values in $[0, \lambda_{\max}]$, with $\E[N_P] = \infty$ when $g^\star = 0$,
\begin{equation}
  \E[N_P] \;\ge\; \frac{\ln(1/\alpha_k)}{g^\star} .
\end{equation}
(b) If every bet equals a fixed $\lambda \in [0, \lambda_{\max}]$ with $g(\lambda) > 0$, then $N_P < \infty$ almost surely and
\begin{equation}
  \frac{\ln(1/\alpha_k)}{g(\lambda)} \;\le\; \E[N_P] \;\le\; \frac{\ln(1/\alpha_k) + \ln(1 + \lambda m_\ell)}{g(\lambda)} .
\end{equation}
\end{lemma}

Part (a) extends the fixed-bet bound of \citet{waudby2025universal} to predictable bets in $[0, \lambda_{\max}]$.

\begin{proof}
Write $Z_j = \ln(1 + \lambda_j (m_\ell - Y_j))$ for the $j$-th increment, so $\ln K_n(m_\ell) = \sum_{j \le n} Z_j$, and note $Z_j \le \ln(1 + \lambda_j m_\ell)$ since $Y_j \ge 0$. (a) The bets are predictable, so given the first $j - 1$ values the bet $\lambda_j$ is fixed and $Z_j$ has conditional mean $g(\lambda_j)$. Hence $M_n = \sum_{j \le n} (Z_j - g(\lambda_j))$ is a martingale, with bounded increments since $Z_j \in [\ln(1/2), \ln(1 + \lambda_{\max} m_\ell)]$. If $\E[N_P] = \infty$ there is nothing to prove. Otherwise optional stopping applies to $M_n$ at $N_P$, so $\E[\sum_{j \le N_P} Z_j] = \E[\sum_{j \le N_P} g(\lambda_j)]$. The left side is at least $\ln(1/\alpha_k) > 0$ by the definition of $N_P$, and $g^\star \ge g(0) = 0$. If $g^\star = 0$ the right side is at most zero, a contradiction, so $\E[N_P] = \infty$. If $g^\star > 0$,
\begin{equation}
\begin{aligned}
  \E[N_P]
  &= \frac{1}{g^\star}\, \E\Big[\sum_{j \le N_P} g^\star\Big]
  && \text{[the sum has $N_P$ terms]} \\
  &\ge \frac{1}{g^\star}\, \E\Big[\sum_{j \le N_P} g(\lambda_j)\Big]
  && \text{[$g(\lambda_j) \le g^\star$ since $\lambda_j \in [0, \lambda_{\max}]$]} \\
  &= \frac{1}{g^\star}\, \E\Big[\sum_{j \le N_P} Z_j\Big]
  && \text{[optional stopping, $\E[M_{N_P}] = 0$]} \\
  &= \frac{1}{g^\star}\, \E\big[\ln K_{N_P}(m_\ell)\big]
  && \text{[$\ln K_n(m_\ell) = \textstyle\sum_{j \le n} Z_j$]} \\
  &\ge \frac{\ln(1/\alpha_k)}{g^\star}
  && \text{[definition of $N_P$]} .
\end{aligned}
\end{equation}
(b) The $Z_j$ are now i.i.d.\ with mean $g(\lambda) > 0$ and values in an interval of width $c_Z = \ln 2 + \ln(1 + \lambda m_\ell)$, since every factor of \cref{eq:capital} is at least $1/2$. For $n > \ln(1/\alpha_k)/g(\lambda)$, Hoeffding's inequality gives $\Prob(N_P > n) \le \Prob(\sum_{j \le n} Z_j < \ln(1/\alpha_k)) \le \exp(-2(n g(\lambda) - \ln(1/\alpha_k))^2/(n c_Z^2))$, which is summable in $n$, so $\E[N_P] = \sum_{n \ge 0} \Prob(N_P > n) < \infty$. Wald's identity then applies to the i.i.d.\ increments $Z_j$ and the stopping time $N_P$. For the lower bound,
\begin{equation}
\begin{aligned}
  \E[N_P]
  &= \frac{1}{g(\lambda)}\, \E\Big[\sum_{j \le N_P} Z_j\Big]
  && \text{[Wald's identity, $\E[Z_j] = g(\lambda)$]} \\
  &= \frac{1}{g(\lambda)}\, \E\big[\ln K_{N_P}(m_\ell)\big]
  && \text{[$\ln K_n(m_\ell) = \textstyle\sum_{j \le n} Z_j$]} \\
  &\ge \frac{\ln(1/\alpha_k)}{g(\lambda)}
  && \text{[definition of $N_P$]} .
\end{aligned}
\end{equation}
For the upper bound, the log-capital is below $\ln(1/\alpha_k)$ one step before promotion, and $\ln K_0(m_\ell) = 0 < \ln(1/\alpha_k)$, so
\begin{equation}
\begin{aligned}
  \E[N_P]
  &= \frac{1}{g(\lambda)}\, \E\Big[\sum_{j \le N_P} Z_j\Big]
  && \text{[Wald's identity, $\E[Z_j] = g(\lambda)$]} \\
  &= \frac{1}{g(\lambda)}\, \E\big[\ln K_{N_P - 1}(m_\ell) + Z_{N_P}\big]
  && \text{[$\ln K_n(m_\ell) = \textstyle\sum_{j \le n} Z_j$]} \\
  &\le \frac{1}{g(\lambda)}\, \big(\ln(1/\alpha_k) + \E[Z_{N_P}]\big)
  && \text{[$\ln K_{N_P - 1}(m_\ell) < \ln(1/\alpha_k)$ by definition of $N_P$]} \\
  &\le \frac{\ln(1/\alpha_k) + \ln(1 + \lambda m_\ell)}{g(\lambda)}
  && \text{[$Z_j \le \ln(1 + \lambda m_\ell)$ since $Y_j \ge 0$]} .
\end{aligned}
\end{equation}
\end{proof}

The proofs use two bounds on the logarithm. For every $x > -1$,
\begin{equation}
  \label{eq:log-upper}
  \ln(1 + x) \;\le\; x - \frac{x^2}{2(1 + \max(x, 0))} ,
\end{equation}
which for $x \le 0$ is Taylor's theorem with a nonpositive third-order remainder, and for $x \ge 0$ is the trapezoid rule applied to the convex function $1/t$ on $[1, 1 + x]$. For every $x \ge -1/2$ \citep{cesabianchi2007improved},
\begin{equation}
  \label{eq:log-lower}
  \ln(1 + x) \;\ge\; x - x^2 .
\end{equation}

\textbf{Proof of the lower bound.} For every $\lambda \in [0, \lambda_{\max}]$,
\begin{equation}
\begin{aligned}
  g(\lambda)
  &= \E\big[\ln\big(1 + \lambda(m_\ell - Y)\big)\big]
  && \text{[definition of $g$]} \\
  &\le \E\left[\lambda(m_\ell - Y) - \frac{\lambda^2 (m_\ell - Y)^2}{2(1 + \lambda m_\ell)}\right]
  && \text{[\cref{eq:log-upper}, $\max(x, 0) \le \lambda m_\ell$]} \\
  &= \lambda \bar s - \frac{\lambda^2 \bar q}{2(1 + \lambda m_\ell)}
  && \text{[$\E[m_\ell - Y] = \bar s$, $\E[(m_\ell - Y)^2] = \bar q$]} \\
  &\le \lambda \bar s - \frac{\lambda^2 \bar q}{2\kappa}
  && \text{[$1 + \lambda m_\ell \le 1 + \lambda_{\max} m_\ell = \kappa$]} .
\end{aligned}
\end{equation}
Dropping the quadratic term gives $g(\lambda) \le \lambda_{\max} \bar s$, and maximizing $\lambda \bar s - \lambda^2 \bar q/(2\kappa)$ over all $\lambda$, at $\lambda = \kappa \bar s/\bar q$, gives $g(\lambda) \le \kappa \bar s^2/(2\bar q)$. Hence
\begin{equation}
\begin{aligned}
  \E[N_P]
  &\ge \frac{\ln(1/\alpha_k)}{g^\star}
  && \text{[\cref{lem:passage}(a)]} \\
  &\ge \ln(1/\alpha_k) \max\left(\frac{2\bar q}{\kappa \bar s^2},\, \frac{1}{\lambda_{\max} \bar s}\right)
  && \text{[$g^\star \le \min(\kappa \bar s^2/(2\bar q),\, \lambda_{\max} \bar s)$]} \\
  &= 2\ln(1/\alpha_k) \max\left(\frac{1}{\kappa}\left(1 + \frac{\sigma_k^2}{s_k^2}\right),\, \frac{\Gamma}{s_k}\right)
  && \text{[\cref{eq:rate-units}]} ,
\end{aligned}
\end{equation}
which is \cref{eq:sample-complexity}.

\textbf{Proof of the upper bound.} In normalized units the bet is $\lambda^\circ = \min(\bar s/(2\bar q), \lambda_{\max})$, by \cref{eq:rate-units}. For every $\lambda \in [0, \lambda_{\max}]$,
\begin{equation}
\begin{aligned}
  g(\lambda)
  &\ge \E\big[\lambda(m_\ell - Y) - \lambda^2 (m_\ell - Y)^2\big]
  && \text{[\cref{eq:log-lower}, $x \ge -1/2$]} \\
  &= \lambda \bar s - \lambda^2 \bar q
  && \text{[$\E[m_\ell - Y] = \bar s$, $\E[(m_\ell - Y)^2] = \bar q$]} .
\end{aligned}
\end{equation}
If $\bar s/(2\bar q) \le \lambda_{\max}$, then $\lambda^\circ = \bar s/(2\bar q)$ and $g(\lambda^\circ) \ge \bar s^2/(2\bar q) - \bar s^2/(4\bar q) = \bar s^2/(4\bar q)$. Otherwise $\lambda^\circ = \lambda_{\max} < \bar s/(2\bar q)$, so $\lambda_{\max}^2 \bar q < \lambda_{\max} \bar s/2$ and $g(\lambda^\circ) \ge \lambda_{\max} \bar s/2$. In both cases $g(\lambda^\circ) \ge \min(\bar s^2/(4\bar q),\, \lambda_{\max} \bar s/2) > 0$. Hence
\begin{equation}
\begin{aligned}
  \E[N_P]
  &\le \frac{\ln(1/\alpha_k) + \ln(1 + \lambda^\circ m_\ell)}{g(\lambda^\circ)}
  && \text{[\cref{lem:passage}(b)]} \\
  &\le \frac{\ln(1/\alpha_k) + \ln\kappa}{g(\lambda^\circ)}
  && \text{[$1 + \lambda^\circ m_\ell \le 1 + \lambda_{\max} m_\ell = \kappa$]} \\
  &\le \big(\ln(1/\alpha_k) + \ln\kappa\big) \max\left(\frac{4\bar q}{\bar s^2},\, \frac{2}{\lambda_{\max} \bar s}\right)
  && \text{[bound on $g(\lambda^\circ)$]} \\
  &= 4\big(\ln(1/\alpha_k) + \ln\kappa\big) \max\left(1 + \frac{\sigma_k^2}{s_k^2},\, \frac{\Gamma}{s_k}\right)
  && \text{[\cref{eq:rate-units}]} ,
\end{aligned}
\end{equation}
which is \cref{eq:cost-upper}. Finally, $N_k = \min(N_P, \Nmax) \le N_P$ by \cref{eq:stopping}.

\textbf{How close the two bounds are.} For $\kappa \ge 1$, $\max(1 + \sigma_k^2/s_k^2,, \Gamma/s_k) \le \kappa \max\big((1 + \sigma_k^2/s_k^2)/\kappa,, \Gamma/s_k\big)$, so the upper bound \cref{eq:cost-upper} exceeds the lower bound \cref{eq:sample-complexity} by at most the factor $2\kappa(1 + \ln\kappa/\ln(1/\alpha_k))$. The Dejavu test does not know $s_k$ and $\sigma_k$, so it cannot place $\lambda^\circ$; its plug-in bets (\cref{app:betting}) are covered by the lower bound, and their cost above it is not analysed here.

\subsection{Proof of \cref{prop:early}}
\label[appendix]{app:early}

Throughout, $m_\ell$ is the normalized threshold of \cref{eq:promote}, $\cF_n$ the history of the first $n$ trajectories, and $g(\lambda) = \E[\ln(1 + \lambda(m_\ell - Y))]$ the growth rate of \cref{sec:rates}, with $g^\star = \max_{\lambda \in [0, \lambda_{\max}]} g(\lambda)$ and $\lambda_{\max}$ the bet cap of \cref{cor:rate}.

The rule uses an upper confidence bound $\hat g_n$ on the maximum growth rate that holds at every buffer size,
\begin{equation}
  \label{eq:ghat-valid}
  \Prob(\exists n : g^\star > \hat g_n) \le \beta .
\end{equation}
We build $\hat g_n$ from betting confidence sequences with the same plug-in bet schedule as the test \citep{waudby2024estimating}, run on the log-capital increments at a grid of fixed bets. By \cref{lem:passage} in \cref{app:rate}, the data a promotion needs is about $\ln(1/\alpha_k)/g^\star$.

Fix $n \le N \le \Nmax$ and work on the event of \cref{eq:ghat-valid}, on which $g^\star \le \hat g_n$. Each bet $\lambda_j$ is predictable, so it is fixed given $\cF_{j-1}$, while $Y_j$ is independent of $\cF_{j-1}$; hence the $j$-th increment of the log-capital, $Z_j = \ln(1 + \lambda_j(m_\ell - Y_j))$, has conditional mean $\E[Z_j \mid \cF_{j-1}] = g(\lambda_j)$. Then
\begin{equation}
\begin{aligned}
  \E\big[\ln K_N(m_\ell) - \ln K_n(m_\ell) \,\big|\, \cF_n\big]
  &= \sum_{j = n+1}^{N} \E\big[Z_j \,\big|\, \cF_n\big]
  && \text{[\cref{eq:capital}]} \\
  &= \sum_{j = n+1}^{N} \E\big[\E[Z_j \mid \cF_{j-1}] \,\big|\, \cF_n\big]
  && \text{[tower property, $\cF_n \subseteq \cF_{j-1}$ for $j > n$]} \\
  &= \sum_{j = n+1}^{N} \E\big[g(\lambda_j) \,\big|\, \cF_n\big]
  && \text{[conditional mean of the increment]} \\
  &\le (N - n)\, g^\star
  && \text{[$\lambda_j \in [0, \lambda_{\max}]$, so $g(\lambda_j) \le g^\star$]} \\
  &\le (N - n)\, \hat g_n
  && \text{[event of \cref{eq:ghat-valid}]} ,
\end{aligned}
\end{equation}
which is \cref{eq:expected-capital}. On the event of \cref{eq:ghat-valid}, $\hat g_n \ge g^\star \ge g(0) = 0$. If the early refusal rule $\ln K_n(m_\ell) + (\Nmax - n)\, \hat g_n < \ln(1/\alpha_k)$ holds at $n$ (\cref{eq:early-refusal}), then for every $N$ with $n \le N \le \Nmax$,
\begin{equation}
\begin{aligned}
  \E\big[\ln K_N(m_\ell) \,\big|\, \cF_n\big]
  &\le \ln K_n(m_\ell) + (N - n)\, \hat g_n
  && \text{[\cref{eq:expected-capital}]} \\
  &\le \ln K_n(m_\ell) + (\Nmax - n)\, \hat g_n
  && \text{[$N \le \Nmax$, $\hat g_n \ge 0$]} \\
  &< \ln(1/\alpha_k)
  && \text{[\cref{eq:early-refusal}]} .
\end{aligned}
\end{equation}
So at every buffer size the test can still reach, its expected log-capital is below the promotion threshold $\ln(1/\alpha_k)$.

\subsection{Proof of \cref{prop:gap}}
\label[appendix]{app:gap}

Throughout, $\cand = \cand_k$ is a fixed candidate. Define its \emph{cost value function} in the production and simulation environments,
\begin{equation}
  V_C^{\cand}(s) = \E_{\tau \sim (\cand, \cM)}\Big[\sum_{t \ge 0} \gc^t c_t \,\Big|\, s_0 = s\Big],
  \qquad
  \tilde V_C^{\cand}(s) = \E_{\tau \sim (\cand, \tilde\cM)}\Big[\sum_{t \ge 0} \gc^t c_t \,\Big|\, s_0 = s\Big],
\end{equation}
so that $J_C(\cand) = \E_{s \sim d_0}[V_C^{\cand}(s)]$ and $\tilde J_C(\cand) = \E_{s \sim \tilde d_0}[\tilde V_C^{\cand}(s)]$. Under \cref{ass:bounded}, both lie in $[0, \cmax / (1 - \gc)]$. The two environments share the cost kernel $\cC$ (\cref{sec:prelim}).

\begin{lemma}[{Simulation lemma for a fixed policy; \citealp{kearns2002near}; \citealp{luo2019algorithmic}}]
\label[lemma]{lem:simlemma}
Define the \emph{one-step dynamics error} $\delta(s, a) = \sum_{s'} \big( T(s' \mid s, a) - \simT(s' \mid s, a) \big)\, V_C^{\cand}(s')$. Then
\begin{equation}
  \label{eq:simlemma-exact}
  J_C(\cand) - \tilde J_C(\cand)
  = \sum_{s} \big( d_0(s) - \tilde d_0(s) \big) V_C^{\cand}(s)
  + \frac{\gc}{1 - \gc}\, \E_{(s, a) \sim \tilde d_{\cand}}\big[ \delta(s, a) \big],
\end{equation}
and consequently $|J_C(\cand) - \tilde J_C(\cand)| \le \varepsilon(\cand)$ with $\varepsilon(\cand)$ as in \cref{eq:gap}.
\end{lemma}

\begin{proof}
Split the difference at the simulation initial distribution:
\begin{equation}
  \label{eq:sim-split}
  J_C(\cand) - \tilde J_C(\cand)
  = \sum_{s} \big( d_0(s) - \tilde d_0(s) \big) V_C^{\cand}(s)
  + \sum_{s} \tilde d_0(s) \big( V_C^{\cand}(s) - \tilde V_C^{\cand}(s) \big).
\end{equation}
The first sum of \cref{eq:sim-split} is the initial-state term of \cref{eq:simlemma-exact}. For the second sum of \cref{eq:sim-split}, the two environments share the cost, and both sums start from $\tilde d_0$, so the telescoping lemma of \citet{luo2019algorithmic}, with the production environment as their $\hat M$ and the simulation environment as their $M$, gives
\begin{equation}
  \sum_{s} \tilde d_0(s) \big( V_C^{\cand}(s) - \tilde V_C^{\cand}(s) \big)
  = \frac{\gc}{1 - \gc}\, \E_{(s, a) \sim \tilde d_{\cand}}\big[ \delta(s, a) \big],
\end{equation}
which proves \cref{eq:simlemma-exact}. For the bound, let $M = \cmax/(1 - \gc)$, so that $V_C^{\cand}$ takes values in $[0, M]$, and use that a function $f$ with values in $[0, M]$ satisfies $|\sum_s (p(s) - q(s)) f(s)| \le M\, D_\mathrm{TV}(p, q)$ for any distributions $p, q$ \citep{levin2017markov}. Then
\begin{equation}
\begin{aligned}
  \big| J_C(\cand) - \tilde J_C(\cand) \big|
  &\le \Big| \sum_{s} \big( d_0(s) - \tilde d_0(s) \big) V_C^{\cand}(s) \Big|
  + \frac{\gc}{1 - \gc}\, \E_{(s, a) \sim \tilde d_{\cand}}\big[ |\delta(s, a)| \big]
  && \text{[\cref{eq:simlemma-exact}]} \\
  &\le M\, \varepsilon_0
  + \frac{\gc M}{1 - \gc}\, \E_{(s, a) \sim \tilde d_{\cand}}\big[ D_\mathrm{TV}\big(T(\cdot \mid s, a), \simT(\cdot \mid s, a)\big) \big]
  && \text{[TV bound, $f = V_C^{\cand}$]} \\
  &= M \left( \varepsilon_0 + \frac{\gc\, \varepsilon_T(\cand)}{1 - \gc} \right)
  && \text{[definition of $\varepsilon_T(\cand)$]} \\
  &= \varepsilon(\cand)
  && \text{[\cref{eq:gap}]} .
\end{aligned}
\end{equation}
\end{proof}

Under \cref{ass:cap}, the proof of \cref{lem:envelope} shows $\E_{\tau \sim (\pi_k, \cM)}[X(\tau)] = J^H_C(\cand_k)$, the candidate's cost return truncated at the horizon. Then
\begin{equation}
  \label{eq:final-slack-bound}
\begin{aligned}
  \ell - J_C(\cand_k)
  &\ge \ell - c_{>H} - J^H_C(\cand_k)
  && \text{[$J_C(\cand_k) - J^H_C(\cand_k) \le c_{>H}$, the truncation offset]} \\
  &= s_k
  && \text{[\cref{eq:slack}, $\E[X(\tau)] = J^H_C(\cand_k)$]} \\
  &\ge \ell - c_{>H} - J_C(\cand_k)
  && \text{[$J^H_C(\cand_k) \le J_C(\cand_k)$, tail costs nonnegative by \cref{ass:bounded}]} \\
  &\ge \ell - c_{>H} - \tilde J_C(\cand_k) - \varepsilon(\cand_k)
  && \text{[\cref{lem:simlemma}]} \\
  &\ge (\ell - \tilde \ell) - c_{>H} - \varepsilon(\cand_k)
  && \text{[simulation constraint, $\tilde J_C(\cand_k) \le \tilde \ell$]} ,
\end{aligned}
\end{equation}
which is \cref{eq:slack-bound}. 

The sim-to-real gap reduces the slack, and so increases the data cost (\cref{cor:rate}), only when the simulation environment is optimistic. An optimistic simulator has $J_C(\cand) > \tilde J_C(\cand)$, and the production slack satisfies $s_k \le \ell - J_C(\cand) < \ell - \tilde J_C(\cand)$ by \cref{eq:final-slack-bound}. When the simulation constraint binds, $\tilde J_C(\cand) = \tilde\ell$, so $s_k \le \ell - \tilde\ell$. The environments of \cref{sec:experiments} are constructed this way.

\section{The Per-Step Clip Alternative}
\label[appendix]{app:perstep}

This appendix explains why Dejavu clips the cumulative importance weight rather than each per-step importance weight, as Retrace \citep{munos2016safe} and V-trace \citep{espeholt2018impala} do. The short version: clipping the cumulative weight discards importance mass only when the whole product drifts above $\Wmax$, and the resulting bias is bounded above by the discarded mass (\cref{prop:clip}). Clipping each per-step weight discards mass at every step where the policies differ, before increments above and below one can offset, and the resulting bias has a lower bound that no amount of production data reduces (\cref{prop:perstep}). The setting and the two identities of \cref{app:clipproof} are used throughout.

\subsection{The per-step alternative}

The per-step clip replaces each factor of the product by $\min(\rho(s_i, a_i), w_{\max})$ for a level $w_{\max} > 0$, giving the statistic
\begin{equation}
\label{eq:xstep}
  X_{\mathrm{step}}(\tau) = \sum_{t<H} \gc^t \Big(\prod_{i \le t} \min\big(\rho(s_i, a_i), w_{\max}\big)\Big)\, c_t .
\end{equation}

Each clipped factor is at most $w_{\max}$, so the clipped product at step $t$ is at most $w_{\max}^{t+1}$ and, with $c_t \le \cmax$,
\begin{equation}
\label{eq:step-envelope}
  X_{\mathrm{step}}(\tau) \le \cmax \sum_{t<H} \gc^t\, w_{\max}^{t+1} = \cmax\, w_{\max}\, \frac{1 - (\gc w_{\max})^H}{1 - \gc w_{\max}} .
\end{equation}
This is the analogue of the range $w_X$ in \cref{eq:certvalue}. It stays bounded as $H$ grows only if $\gc w_{\max} < 1$, that is, $w_{\max} < 1/\gc$. Above that level it grows like $(\gc w_{\max})^H$, and since the headroom $\Gamma$, which grows with the range, enters the range term of both bounds of \cref{cor:rate} linearly, such a level is unusable at the horizons of \cref{sec:experiments}. The admissible per-step clip levels are therefore $w_{\max} \in (0, 1/\gc)$.

\begin{proposition}[Per-step clip bias lower bound]
\label[proposition]{prop:perstep}
Under the conditions of \cref{prop:clip}, for any $w_{\max} \in (0, 1/\gc)$,
\begin{equation}
\label{eq:step-floor}
  J^H_C(\cand) - \E_{\tau \sim (\pi, \cM)}[X_{\mathrm{step}}(\tau)] \;\ge\; J^H_C(\cand) - w_{\max}\, J^H_C(\pi; \gc w_{\max}) .
\end{equation}
\end{proposition}

\textbf{Proof.} By the same factor bound as in \cref{eq:step-envelope}, but keeping the cost, $X_{\mathrm{step}}(\tau) \le \sum_{t<H} \gc^t w_{\max}^{t+1} c_t$ on every trajectory, since $c_t \ge 0$. Taking expectations,
\begin{equation}
  \E_{\tau \sim (\pi, \cM)}[X_{\mathrm{step}}(\tau)] \le w_{\max} \sum_{t<H} (\gc w_{\max})^t\, \E_{\tau \sim (\pi, \cM)}[c_t] = w_{\max}\, J^H_C(\pi; \gc w_{\max}) ,
\end{equation}
Subtracting from $J^H_C(\cand)$ gives \cref{eq:step-floor}.

The right-hand side of \cref{eq:step-floor} does not involve the buffer, so no amount of production data reduces it. It is informative when the candidate is costlier than the baseline policy, which is exactly the case a test exists to detect. At $w_{\max} = 1$, the level used by Retrace and V-trace, the right-hand side is $J^H_C(\cand) - J^H_C(\pi)$ for every $\gc$: the per-step-clipped statistic cannot see any of the cost the candidate adds over the baseline policy. Levels of $w_{\max}$ above one recover some of it, but admissibility forces $w_{\max} \to 1$ as $\gc \to 1$. In contrast, the bound of \cref{prop:clip} on the cumulative clip is zero whenever the retained mass is one, whatever the candidate's cost.

\subsection{Why the two clips behave so differently}

Write the cumulative importance weight in log space as the random walk
\begin{equation}
  S_t = \ln \cW(\tau) = \sum_{i \le t} Z_i, \qquad Z_i = \ln \rho(s_i, a_i) .
\end{equation}
At any state where the policies differ, $\rho(s, a)$ is a nonconstant nonnegative random variable over $a \sim \pi(\cdot \mid s)$ with mean one by \cref{eq:unit-mean}. So it takes values on both sides of one, $\Prob(\rho(s, a) > 1) > 0$, and by Jensen's inequality its logarithm has negative mean, $\E_{a \sim \pi(\cdot \mid s)}[Z] = -D_{\mathrm{KL}}(\pi \,\|\, \cand) < 0$. The walk $S_t$ therefore drifts downward on average, with increments that are positive on a positive fraction of steps.

The cumulative clip acts on the sum, $\min(S_t, \ln \Wmax)$. Positive and negative increments offset inside $S_t$ before the clip is applied, and the clip binds only if the walk climbs all the way from its downward drift to $\ln \Wmax$. With a small anchor this requires a long run of positive increments, which is why the anchor of \cref{eq:proposer} makes \cref{ass:cap} hold in practice, and why the mass it discards can be audited through $\varrho_t$.

The per-step clip acts on each increment, $\min(Z_i, \ln w_{\max})$, before any offsetting. Since
\begin{equation}
  \E_{a \sim \pi(\cdot \mid s)}\big[\min(Z, \ln w_{\max})\big] = \E_{a \sim \pi(\cdot \mid s)}[Z] - \E_{a \sim \pi(\cdot \mid s)}\big[(Z - \ln w_{\max})^+\big],
\end{equation}
every step adds a nonnegative leak $\E_{a \sim \pi(\cdot \mid s)}[(Z - \ln w_{\max})^+]$ to the downward drift, strictly positive at every state where $\Prob(\rho(s, a) > w_{\max}) > 0$. The leaks accumulate over steps and are never recovered, and the mass they remove is what \cref{prop:perstep} measures. The leak is large whenever the spread of $Z$ is large relative to $\ln w_{\max}$, which the next paragraph quantifies.

Fix a state $s$, take the action space discrete so that sums are finite, and write $\cand(a) = \pi(a) + \delta(a)$ with $\delta$ small and $\sum_a \delta(a) = 0$, since both are probability distributions. Then $Z = \ln(1 + \delta(a) / \pi(a))$ for $a \sim \pi$. Using $\ln(1 + x) \approx x - x^2/2$ and dropping terms of order $\delta^3$,
\begin{equation}
  \E_{\pi}[Z] \approx \sum_a \pi(a) \Big(\frac{\delta(a)}{\pi(a)} - \frac{\delta(a)^2}{2\pi(a)^2}\Big) = -\frac{1}{2} \sum_a \frac{\delta(a)^2}{\pi(a)},
  \qquad
  \E_{\pi}[Z^2] \approx \sum_a \pi(a) \frac{\delta(a)^2}{\pi(a)^2} = \sum_a \frac{\delta(a)^2}{\pi(a)} .
\end{equation}
Since $\E_\pi[Z]^2$ is of order $\delta^4$, $\mathrm{Var}_\pi(Z) \approx \sum_a \delta(a)^2 / \pi(a)$. The same expansion of the KL divergence gives
\begin{equation}
  D_{\mathrm{KL}}(\cand \,\|\, \pi) = \sum_a \big(\pi(a) + \delta(a)\big) \ln\Big(1 + \frac{\delta(a)}{\pi(a)}\Big)
  \approx \sum_a \delta(a) + \sum_a \frac{\delta(a)^2}{\pi(a)} - \frac{1}{2} \sum_a \frac{\delta(a)^2}{\pi(a)}
  = \frac{1}{2} \sum_a \frac{\delta(a)^2}{\pi(a)},
\end{equation}
so that, to second order in $\delta$,
\begin{equation}
\label{eq:var-2kl}
  \mathrm{Var}_\pi(Z) \approx 2\, D_{\mathrm{KL}}(\cand \,\|\, \pi) .
\end{equation}
Both KL directions equal $\frac{1}{2}\sum_a \delta(a)^2/\pi(a)$ at this order, so the exact mean $\E_{a \sim \pi(\cdot \mid s)}[Z] = -D_{\mathrm{KL}}(\pi \,\|\, \cand)$ is also $\approx -\frac{1}{2}\sum_a \delta(a)^2/\pi(a)$. At a state with divergence $D_{\mathrm{KL}}$, the per-step log-weight therefore has mean $-D_{\mathrm{KL}}$ and standard deviation $\sqrt{2 D_{\mathrm{KL}}}$. For small divergences the standard deviation is far larger than the absolute value of the mean, so a positive draw of $Z$, on which the per-step clip binds once it exceeds $\ln w_{\max}$, is ordinary fluctuation rather than a rare event.

\textbf{The example $\gc = 0.99$.} Admissible levels then satisfy $\ln w_{\max} < \ln(1/0.99) \approx 0.01$, and any draw of $Z$ above $\ln w_{\max}$ is clipped by the per-step clip. By \cref{eq:var-2kl} the standard deviation of $Z$ exceeds $0.01$ as soon as $D_{\mathrm{KL}}(\cand \,\|\, \pi) > 5 \times 10^{-5}$ at that state, a divergence far too small to allow any useful policy change. Take $D_{\mathrm{KL}} = 10^{-3}$. Then $Z$ has mean $-0.001$ and standard deviation $0.045$, so a draw less than a quarter of a standard deviation above its mean is already clipped. Each clip removes $Z - \ln w_{\max}$ from the log-weight, and the removal is permanent: later negative increments are added to the clipped value, not to the original. This is the leak of the previous paragraph, it recurs at every state where the policies differ, and its accumulation is what the lower bound of \cref{prop:perstep} measures. The cumulative clip sees the same increments but clips only the running sum $S_t$, and only once $S_t$ exceeds $\ln \Wmax \ge 0.5$, fifty times the admissible per-step level $\ln(1/\gc)$. No single step can reach that: the sum has to climb through many positive increments that negative ones did not cancel, and whether it ever does depends on the drift and the horizon of the whole walk. That is what the retained mass $\varrho_t$ can measure.

\section{The Doubly-Robust Statistic}
\label[appendix]{app:dr}

Dejavu computes its statistic $X(\tau)$ by importance weighting alone in \cref{eq:certvalue}. The doubly-robust estimator \citep{jiang2016doubly, thomas2016data} is the natural alternative: it subtracts a cost model trained in simulation as a control variate and can have a lower variance than the plain importance-weighted estimator. Let $\hat{c}_t = \hat{c}(s_t, a_t)$ be a per-step cost estimator trained in the simulation environment, with residual $d_t = c_t - \hat{c}_t$ and candidate expectation $\bar{c}_t = \E_{a \sim \cand_k(\cdot \mid s_t)}[\hat{c}(s_t, a)]$. Clipping and truncating the estimator as in \cref{eq:certvalue} gives the \emph{doubly-robust statistic}
\begin{equation}
  \label{eq:dr}
  X_{\mathrm{DR}}(\tau)=\sum_{t=0}^{H-1}\gc^t\Big[\bar{\rho}_{0:t}(\tau)\,d_t+\bar{\rho}_{0:t-1}(\tau)\,\bar{c}_t\Big],
\end{equation}
with $\bar{\rho}_{0:-1} \equiv 1$, whose values lie in $[-w_X, 2w_X]$ rather than $[0, w_X]$ (shown below). This raises the headroom $\Gamma$ of \cref{cor:rate} from $w_X - \ell + c_{>H}$ to $2w_X - \ell + c_{>H}$, at least double since $\ell - c_{>H} \ge s_k > 0$, which a cost model cannot reduce, whereas the variance term it can. A poor cost model does not bias the doubly-robust statistic but can raise its variance above that of $X(\tau)$, which is the member of the same family with $\hat{c} \equiv 0$. The statistic a test runs on must be fixed before its buffer is collected (\cref{ass:indep}), so the doubly-robust form cannot be adopted only on tests where its model helps; its enlarged range would be paid on every test.

Let the cost model take values in $[0, \cmax]$, as the costs do (\cref{ass:bounded}). Then for every step $t < H$,
\begin{equation}
  d_t = c_t - \hat{c}_t \in [-\cmax, \cmax],
  \qquad
  \bar{c}_t = \E_{a \sim \cand_k(\cdot \mid s_t)}[\hat{c}(s_t, a)] \in [0, \cmax],
  \qquad
  \cWc, \bar{\rho}_{0:t-1} \in [0, \Wmax],
\end{equation}
where the last pair holds because each clipped cumulative weight is a product of nonnegative factors capped at $\Wmax$. Multiplying,
\begin{equation}
  \cWc\, d_t \in [-\Wmax \cmax,\; \Wmax \cmax],
  \qquad
  \bar{\rho}_{0:t-1}\, \bar{c}_t \in [0,\; \Wmax \cmax],
\end{equation}
so each summand of \cref{eq:dr} lies in $[-\Wmax \cmax,\; 2 \Wmax \cmax]$. Summing the discount series,
\begin{equation}
  -\Wmax \cmax \sum_{t=0}^{H-1} \gc^t
  \;\le\; X_{\mathrm{DR}}(\tau)
  \;\le\; 2\, \Wmax \cmax \sum_{t=0}^{H-1} \gc^t,
  \qquad
  \Wmax \cmax \sum_{t=0}^{H-1} \gc^t = \Wmax \cmax\, \frac{1 - \gc^H}{1 - \gc} = w_X,
\end{equation}
which is $X_{\mathrm{DR}}(\tau) \in [-w_X, 2w_X]$, an interval of width $3w_X$ against the width $w_X$ of $X(\tau) \in [0, w_X]$ in \cref{eq:certvalue}. The betting confidence sequence of \cref{sec:prelim} requires values in $[0, 1]$, so a test run on the doubly-robust statistic would normalize by $(X_{\mathrm{DR}} + w_X) / (3w_X)$, and \cref{cor:rate} holds with the range width $3w_X$ in place of $w_X$ and the headroom $2w_X - \ell + c_{>H}$ in place of $\Gamma$.

\section{Bet Schedule and Computation}
\label[appendix]{app:betting}

For reproducibility we record the bet schedule and the search procedure. We use the predictable plug-in schedule of \citet{waudby2024estimating}, with running mean and variance estimates
\begin{equation}
  \hat\mu_j = \frac{1/2 + \sum_{i \le j} Y_i}{j + 1},
  \qquad
  \hat\sigma_j^2 = \frac{1/4 + \sum_{i \le j} (Y_i - \hat\mu_i)^2}{j + 1},
\end{equation}
the bet at sample $j$ is
\begin{equation}
  \lambda_j(m) = \min\left( \sqrt{\frac{2 \ln(1/\alpha_k)} {\hat\sigma_{j-1}^2 \, j \ln(j+1)}},\; \frac{1}{2(1-m)} \right).
\end{equation}
$K_n(m)$ is monotone in $m$, so $U_n(\cand_k)$ is computed by binary search on $m$, evaluating $\sum_j \log(1 + \lambda_j(m) (m - Y_j))$ in log space; $40$ iterations resolve $m$ beyond single-precision limits. Because validity holds for any predictable schedule, none of these implementation choices affect the guarantees, only tightness.

\section{Experimental Details}
\label[appendix]{app:experiments}
\providecommand{\tbd}{\textcolor{red}{TBD}}

Every result uses 20 seeds. For each seed, every method starts from the same initial policy and faces the same draw of the production environment. \Cref{tab:shared,tab:env-settings,tab:hyper} give the settings and \cref{tab:prod-data} the production data consumed.

\begin{table}[ht]
\centering
\caption{Settings shared by every method.}
\label{tab:shared}
\footnotesize
\begin{tabular}{ll}
\toprule
\textbf{Setting} & \textbf{Value} \\
\midrule
Seeds & $20$ \\
Root seed & $0$ ($1$ for CPO's production training) \\
Prod param seed & $101$, $202$, $0$ (HalfCheetahVelocity, PointGoal, IcyLake) \\
Policy & diagonal Gaussian (HalfCheetahVelocity, PointGoal); categorical (IcyLake) \\
Networks & actor, reward critic and cost critic, each $2 \times 256$, $\tanh$ \\
Optimizer & Adam, $\epsilon = 10^{-5}$, learning rate $3 \times 10^{-4}$, gradient-norm clip $0.5$ \\
Reward discount $\gamma$ & $0.99$ \\
Cost discount $\gc$ & $0.99$ \\
GAE parameter & $0.95$ \\
\bottomrule
\end{tabular}
\end{table}

\begin{table}[ht]
\centering
\caption{Environment settings, identical across methods. Production parameters are drawn once per seed, uniformly over the range shown. $^\ast$Selected among $\tilde\ell \in \{1, 0.5\}$ (HalfCheetahVelocity, warm-started from an unconstrained policy) or $\{2, 1\}$ (PointGoal) by a $64$-episode evaluation at the most adverse corner of the production range.}
\label{tab:env-settings}
\footnotesize
\setlength{\tabcolsep}{3pt}
\begin{tabular}{lccc}
\toprule
\textbf{Setting} & \textbf{HalfCheetahVelocity} & \textbf{PointGoal} & \textbf{IcyLake} \\
\midrule
Cost trigger & speed $> 3.21$\,m/s & in a hazard or by a wall & on an icy tile \\
Cost limit $\ell$ & $5$ & $8$ & $1.4$ \\
Simulation cost limit $\tilde\ell$ & $5$ & $8$ & $1.4$ \\
Cost bound $\cmax$ & $1$ & $1$ & $1$ \\
Episode length & $1000$ & $1000$ & $100$ \\
Horizon $H$ & $1000$ & $1000$ & $100$ \\
Simulation environment & gain, mass, damping $1$ & gain, mass, damping $1$ & success $0.50$ \\
Production gain & $[1.1, 1.3]$ & $[1.1, 1.3]$ & -- \\
Production mass & $[0.9, 1.0]$ & $[0.9, 1.05]$ & -- \\
Production damping & $[0.8, 1.0]$ & $[0.65, 0.9]$ & -- \\
Production actuator lag & none & one step (previous action applied) & -- \\
Production success & -- & -- & $[0.30, 0.40]$ \\
Initial policy & CRPO in simulation, $\tilde\ell = 0.5$$^\ast$ & CRPO in simulation, $\tilde\ell = 1$$^\ast$ & fixed, biased to one action \\
\bottomrule
\end{tabular}
\end{table}

\begin{table}[ht]
\centering
\caption{Per-method settings. Simulation training (rows marked $^\dagger$) is shared by the Dejavu proposer and SPiDR.}
\label{tab:hyper}
\footnotesize
\setlength{\tabcolsep}{3pt}
\begin{tabular}{llccc}
\toprule
\textbf{Method} & \textbf{Setting} & \textbf{HalfCheetahVelocity} & \textbf{PointGoal} & \textbf{IcyLake} \\
\midrule
\multirow{18}{*}{Dejavu}
 & Tests (maximum) & $50$ & $50$ & $40$ \\
 & Simulation steps per test$^\dagger$ & $4.1$M & $4.1$M & $4.1$M \\
 & Parallel simulation environments$^\dagger$ & $256$ & $256$ & $256$ \\
 & Rollout length, epochs, minibatch$^\dagger$ & $128$, $4$, $512$ & $128$, $4$, $512$ & $128$, $4$, $512$ \\
 & Ratio clip, reward and cost$^\dagger$ & $0.2$, $0.05$ & $0.2$, $0.05$ & $0.2$, $0.05$ \\
 & Value-loss coefficient$^\dagger$ & $0.5$ & $0.5$ & $0.5$ \\
 & Entropy coefficient$^\dagger$ & $0.0075$ & $0.0075$ & $0.02$ \\
 & Anchor coefficient (\cref{eq:proposer}) & $2$ & $2$ & $2$ \\
 & Clip level $\ln \Wmax$ & $1.0$ & $1.0$ & $1.0$ \\
 & Chunk size $n_0$ & $32$ & $32$ & $64$ \\
 & Data budget per test $\Nmax$ & $1024$ & $1024$ & $4096$ \\
 & Production budget (steps) & $10$M & $10$M & $1$M \\
 & Initial-policy trajectories & $256$ & $256$ & chunks of $64$ until it passes, at most $1024$ \\
 & Lifetime budget $\alpha$, initial $\alpha_{\pi_0}$ & $0.075$, $0.025$ & $0.075$, $0.025$ & $0.075$, $0.025$ \\
 & Early-refusal level (\cref{eq:ghat-valid}) & $0.1$ & $0.1$ & $0.1$ \\
 & Early refusal checked from & $256$ trajectories & $256$ trajectories & $256$ trajectories \\
 & Retained mass (mean) & $0.80$ & $0.92$ & $0.98$ \\
\midrule
\multirow{6}{*}{SPiDR}
 & Pessimism coefficient $\lambda$ & $0.1$ & $0.1$ & $5$ \\
 & Ensemble size & $8$ & $8$ & $8$ \\
 & Randomized parameter & gain, $[1.0, 1.4]$ & gain, $[1.0, 1.4]$ & success, $[0.25, 0.55]$ \\
 & Anchor coefficient & $4$ & $4$ & $2$ \\
 & Tests & $30$ & $30$ & $10$ \\
\midrule
\multirow{6}{*}{CPO}
 & Production environments, rollout & $64$, $128$ & $64$, $128$ & $64$, $128$ \\
 & Production budget (steps) & $10$M & $10$M & $1.05$M \\
 & Trust region & $0.01$ & $0.01$ & $0.01$ \\
 & Backtracking (coefficient, steps) & $0.8$, $10$ & $0.8$, $10$ & $0.8$, $10$ \\
 & CG damping, full-batch critic steps & $0.1$, $40$ & $0.1$, $40$ & $0.1$, $40$ \\
 & Evaluation freq & $25$ updates & $25$ updates & $8$ updates \\
\midrule
\multirow{5}{*}{HCPI}
 & Episodes per iteration (train, test) & $320$ ($64$, $256$) & $320$ ($64$, $256$) & $320$ ($64$, $256$) \\
 & Test level $\delta$, per test & $0.075$ & $0.075$ & $0.075$ \\
 & CPO trust region, step sizes & $0.01$, $\{1, \tfrac12, \tfrac14, \tfrac18\}$ & $0.01$, $\{1, \tfrac12, \tfrac14, \tfrac18\}$ & $0.01$, $\{1, \tfrac12, \tfrac14, \tfrac18\}$ \\
 & Production budget (steps) & $10$M & $10$M & $1$M \\
\bottomrule
\end{tabular}
\end{table}

\begin{table}[ht]
\centering
\caption{Production steps consumed per seed, mean (maximum) over seeds. Dejavu includes the initial policy's certification. Tuning trials and truth evaluations are not counted for any method.}
\label{tab:prod-data}
\footnotesize
\begin{tabular}{lccc}
\toprule
\textbf{Method} & \textbf{HalfCheetahVelocity} & \textbf{PointGoal} & \textbf{IcyLake} \\
\midrule
Dejavu & $9.57$M ($9.98$M) & $8.65$M ($9.98$M) & $0.80$M ($1.00$M) \\
SPiDR & $0$ & $0$ & $0$ \\
CPO & $10$M & $10$M & $1$M \\
HCPI & $9.86$M ($9.86$M) & $9.86$M ($9.86$M) & $0.98$M ($1.00$M) \\
\bottomrule
\end{tabular}
\end{table}

\paragraph{Ground truth evaluation.}
We estimate a policy's production cost return $J_C(\pi)$ by $\hat J_C(\pi)$, the mean discounted cost return of held-out production episodes of $\pi$. On HalfCheetahVelocity and PointGoal we run $256$ episodes of $1000$ steps, and add batches of $256$ episodes, up to $2048$, while $\hat J_C(\pi)$ is within two standard errors of $\ell$, so that policies near the limit are classified as safe or unsafe with more episodes. On IcyLake $1024$ environments each run $1000$ steps with automatic resets, giving at least $10{,}240$ episodes. Across methods, every policy deployed to production is evaluated this way, apart from CPO which is evaluated at regular update intervals defined in \cref{tab:hyper}.

\paragraph{Tuning budget.}
A default setting for Dejavu's anchor coefficient and clip level were chosen by results on an external lightweight benchmark not including the CRAX environments, and are applied in \cref{tab:hyper}. The other settings were chosen based on existing sensible defaults on other popular RL methods and the environment data constraints. CPO's trust-region settings are those of the reference implementation of \citet{ray2019benchmarking}, and its networks, critic steps and batch size are the shared settings of \cref{tab:shared,tab:hyper}. HCPI keeps the train split, per-test level and test-set handling of \citet{thomas2015improvement}, with one CPO step on the training episodes as its candidate search. SPiDR's \emph{pessimism coefficient} $\lambda$, the weight of the penalty in its penalized cost, follows \citet[Appendix~E]{as2025spidr}, with one added condition. Its starting value $\lambda_0$ makes the penalty, whose mean per step is estimated in the simulation environment, about as large as the largest per-step cost, which is $1$ in every environment: $\lambda_0 = 7$, $0.66$ and $120$ on HalfCheetahVelocity, PointGoal and IcyLake. Then $\lambda$ is lowered over about five values, each run on $3$ production seeds, and we keep the value with the highest final reward among those with the fewest seeds ever deploying a policy with $\hat J_C(\pi) > \ell$. The added condition is that a value of $\lambda$ is used only if some policy keeps its discounted cost plus $\lambda$ times its discounted penalty within $\ell$ in the simulation environment. Otherwise SPiDR's constraint cannot be met, and its constrained optimizer (CRPO \citet{xu2021crpo}) only takes steps that lower the penalized cost, never steps on the reward. The values of $\lambda$ were $\{7, 0.7, 0.1, 0.03, 0\}$ on HalfCheetahVelocity, $\{0.1, 0.03, 0.01, 0\}$ on PointGoal (starting below $\lambda_0$) and $\{10, 5, 3, 1\}$ on IcyLake ($\lambda_0$ fails the condition), giving $0.1$, $0.1$ and $5$ (\cref{tab:hyper}).

\end{document}